\documentclass{bmcart}

\usepackage{amsthm,amsmath}
\usepackage{siunitx}
\RequirePackage{hyperref}
\usepackage[utf8]{inputenc} 
\usepackage{graphicx}
\usepackage{subcaption}
\usepackage{tablefootnote}

\startlocaldefs
\endlocaldefs

\begin{document}

\begin{frontmatter}

\begin{fmbox}
\dochead{Regular Article}


\title{Does Noise Affect Housing Prices? A Case Study in the Urban Area of Thessaloniki}


\author[
   addressref={aff1},                   
   corref={aff1},                       
   email={georgiok@csd.auth.gr}   
]{\inits{GK}\fnm{Georgios} \snm{Kamtziridis}}
\author[
   addressref={aff1},
   email={dvrakas@csd.auth.gr}
]{\inits{DV}\fnm{Dimitris} \snm{Vrakas}}
\author[
   addressref={aff1},
   email={greg@csd.auth.gr}
]{\inits{GT}\fnm{Grigorios} \snm{Tsoumakas}}


\address[id=aff1]{
  \orgname{Department of Informatics, Aristotle University of Thessaloniki}, 
  \city{Thessaloniki},                              
  \cny{GR}                                    
}
\address[id=aff2]{%
  \orgname{Department of Informatics, Aristotle University of Thessaloniki},
  \city{Thessaloniki},                              
  \cny{GR}    
}


\begin{artnotes}
\end{artnotes}

\end{fmbox}


\begin{abstractbox}

\begin{abstract} 
Real estate markets depend on various methods to predict housing prices, including models that have been trained on datasets of residential or commercial properties. Most studies endeavor to create more accurate machine learning models by utilizing data such as basic property characteristics as well as urban features like distances from amenities and road accessibility. Even though environmental factors like noise pollution can potentially affect prices, the research around this topic is limited. One of the reasons is the lack of data. In this paper, we reconstruct and make publicly available a general purpose noise pollution dataset based on published studies conducted by the Hellenic Ministry of Environment and Energy for the city of Thessaloniki, Greece. Then, we train ensemble machine learning models, like XGBoost, on property data for different areas of Thessaloniki to investigate the way noise influences prices through interpretability evaluation techniques. Our study provides a new noise pollution dataset that not only demonstrates the impact noise has on housing prices, but also indicates that the influence of noise on prices significantly varies among different areas of the same city.
\end{abstract}


\begin{keyword}
\kwd{Housing Prices Prediction}
\kwd{Noise Pollution}
\kwd{Ensemble Models}
\kwd{Interpretability}
\end{keyword}


\end{abstractbox}
%

\end{frontmatter}



\section*{Introduction}
The real estate market plays an important role in people's lives, from individuals and families, to small businesses and large corporations. The process of purchasing or renting a property, whether for residential or commercial purposes, mainly depends on the economic and financial planning of a family or a company. Additionally, it is strongly related to the macroeconomics and the financial stability of much larger groups of people such as countries. Any sign of inconsistency or fluctuation in the real estate market can provoke apprehension in the state, trigger an economic recession or, ultimately, even lead to financial crises through \textit{housing bubble bursts}. The potential risks are well known to the concerned parties and more importantly to governments that monitor the market on a regular basis. Banks have also invested greatly in real estate in order to obtain accurate house pricing estimates for mortgages and housing loans. These organizations often need to estimate the value of a given property for auctions or damage control when clients are unable to pay their debts. Besides states and organizations, property owners and investors should have the right to access valuable insights about the value of their properties too. This knowledge can increase the efficiency of managing assets or even help make profitable property investments.

Property estimations are performed by human experts like real estate brokers and engineers. This estimation process considers properties’ features and amenities, as well as external factors such as bus station density or distances to city centers. These are combined with other metrics, like the House Price Index~\cite{improved_machine_learning_techniques}, which tracks the changes in property prices, to arrive at a price estimate. During this process, there is no way of quantifying the accuracy of prediction nor the importance of each component that was included in the task. Therefore, the absence of confidence increases the risk of the forthcoming decision, which can end up being financially harmful.

In the contemporary world, the real estate market is represented mostly through different web-based services. In each country, there are numerous websites with vast amounts of properties available for renting or buying. These data have been utilized in the past for different analyses, ranging from creating models capable of predicting house prices based on their features to estimating prices over time in order to understand their seasonality. There has been a lot of research on this topic over the years, with big real estate datasets containing hundreds of properties being used to train machine learning models with the ultimate goal of providing meaningful price estimates. These datasets contain basic property features that are specific to the building itself, such as location, size, floor level and heating type to name a few. Moreover, they can incorporate other features related to the surrounding area of the property, such as road network accessibility and distances from basic points of interest. All these features contribute to the urban profile of a neighborhood, which can directly or indirectly affect prices to a great extent. The importance of these features and their correlation to the price estimates have been validated in previous research~\cite{economic_value_neighborhoods,Baldominos_2018, park_2015,truong_2019}.

Environmental factors have not been taken into consideration in the literature as much as they should have, despite their obvious role when selecting a property. The two most popular are the air quality index and the noise pollution. The first indicates the level of cleanliness in the air that influences the overall health of the population in a given area. The second one is related to the actual noise caused by road traffic, crowds, aviation and other factors such as the presence of night clubs or manufacturing establishments. Although these two features play a major role in the nature of a neighborhood, research on their impact on house prices remains largely underexplored. To some extent, this is to be expected given the practical challenges of gathering environmental data, such as expensive measuring and monitoring tools, specialized software, and on-site orchestration of distributed sensors. Usually, these studies are conducted by large corporations or state departments that subsequently hold the data for internal use. Even when the results are made public, the raw data that were collected and used in the experiments are typically not made available.

The impact of the real estate market on a country, in addition to the innovations that can emerge through research in the field, highlights the potential profit of such work. Being able to generate valuable environmental features of an urban area and, then, use those in the housing price prediction problem can help individuals, small and medium-sized businesses, all the way to large corporations, banks and government experts make profitable decisions. Aside from profitability, it can shed light on the various factors that influence prices. Knowing if and how the environment affects housing prices can assist urban planners to design more functional and efficient cities.

In the first part of this paper, we extract environmental data, and more specifically noise pollution, from published scientific studies. We focus on studies performed by the Hellenic Ministry of Environment and Energy\footnote{\url{https://ypen.gov.gr/}} for the urban area of Thessaloniki, Greece. The end results were published by the government with heat maps demonstrating the spatial distribution of noise across the city. However, none of the core noise measurements were made public, making any future use or contribution to the field difficult. We have managed to overcome this limitation by meticulously re-creating the sense of noise into a general-purpose and easy to use dataset.

In the second part of this work, we highlight the importance of noise in predicting house prices. To verify this, we have used the property database of Openhouse\footnote{\url{https://openhouse.gr}}, which is a real estate platform operating in major cities of Greece and, mainly, in the area of Thessaloniki. Since we need to create accurate models we choose to use \textit{ensemble} methods that proved to work well in the research literature. The property and the noise data are used to create multiple models with distinct configurations, exploring different aspects of the same problem. 

The main contributions of this work are:
\begin{enumerate}
    \item A new general-purpose sense-of-noise dataset, as well as a new housing price dataset containing noise information for the area of Thessaloniki\footnote{\url{https://github.com/gkamtzir/housing-prices-and-noise-thessaloniki}}.
    \item An extensive experimental evaluation of the contribution of noise in the property price estimation process via ensemble models such as \textit{XGBoost}~\cite{chen_2016} and \textit{light gradient boosting}~\cite{ke_2017} models.
\end{enumerate}

\section*{Related Work}
This section presents relevant research in the field of housing price prediction from a data perspective. It is important to discuss key relevant work in order to better understand the current state of the area, as well as to position this paper properly within the literature. We begin by outlining the basic solutions proposed along the years on housing price estimates that take basic property features into consideration. Then, we move to more advanced approaches, which incorporate various environmental information and, eventually, focus on experiments based on noise pollution.

Baldominos~\cite{Baldominos_2018} studies the housing price prediction problem in the Salamanca district of Madrid. With a collection of 2,266 properties from popular online sites containing the fundamental characteristics, they test the correlation between the features and the price to find out that size is the most important one. They use these data to construct various regression models of different specifications, such as support vector machines, multi-layer perceptrons and ensembles of regression trees, all trying to predict prices given the features. The final results showcase the superiority of the ensemble trees when compared to others. Imran~\cite{house_price_islamabad} follows another approach for the capital of Pakistan, Islamabad. Alongside the basic property characteristics, they gather some features related to the surrounding area of a property. For instance, they attempt to include neighborhood related information through binary values (yes/no) indicating the existence of core amenities and services like hospitals, schools and entertainment. Although their experiments encapsulate many features, the results show that besides the total size, the number of bedrooms and bathrooms, also, radically influence the price, with support vector machines being the best performing model.

Truong~\cite{truong_2019} focuses on the Beijing area by using the “Housing Price in Beijing” dataset which contains more than 300,000 properties. Each property, apart from its standard attributes, has various spatial information like distance from the city center and subway accessibility. The exploratory analysis demonstrates direct correlation between the location and the property price, since each district has a different price range. Initially, random forest~\cite{random_forests}, XGBoost and lightweight gradient boosting models were used for training. Then, the authors combine these to build a stacked generalization model~\cite{wolpert} by placing random forest and lightweight gradient boosting at the first level and XGBoost at the second one. This architecture outperforms any of the individual ones in terms of accuracy, with a much higher computational cost. Similarly, Xue~\cite{xue_2020} accumulates property data and urban details like bus and metro stations and routes, traffic and road network information for the city of Xi’an, China. The urban data are preprocessed and new meaningful indices are introduced. The property features and the new indices are utilized by ensemble models to highlight the fact that size is, again, the most influential factor in the matter of predicting prices. Additionally, they illustrate the importance of the neighborhood of a property, because the next most important group of features is related to the spatial indices. Along the same lines, Kang~\cite{kang_2020} engineers relevant features from more generic urban characteristics like human mobility patterns and socioeconomic data. They experiment with a gradient boosting ensemble in order to analyze features’ significance, where they come to the conclusion that some spatial features can play a more decisive role when it comes to predicting prices. For example, the prices of properties located near university campuses are mainly affected by the distance to the campus rather than their total size.

Environmental conditions can, also, act on prices. Chiarazzo~\cite{chiarazzo_2014} gathers property and air pollution data for the city of Taranto in Italy, which is marked as a high environmental risk area due to its heavy industry. With feature selection and an artificial neural network they put to the test the correlation of each feature through an one-by-one elimination process. Interestingly, they state that sulfur dioxide concentration, one of the five major air pollutants, is the most determinant with respect to price, ranking higher than other characteristics such as floor level and distance to the city center. Shanghai is another industrialized city, where Zou~\cite{zou_2022} evaluates the air pollution phenomenon in connection with property prices to quantify even more their relation. A total of 27,608 properties in conjunction with air pollutants are used as training data in a gradient boosting model which it attributes 1.6\% in terms of contribution. Under no circumstance, this percentage can be considered as minimal, since a reduction of 1 \si{\micro\gram}/$m^3$ in nitrogen dioxide increases the price by roughly 278 Yuan per square meter.

Regarding noise pollution, there is much less research available attempting to correlate house prices to noise levels. In general, noise pollution is measured in decibels, where higher values suggest noisier environments. Blanco~\cite{blanco_2011} uses hedonic models to analyze the connection between prices and noise levels in three different areas in the United Kingdom. They suggest that when evaluating properties with similar amenities the presence or absence of noise affects people’s choices. In particular, the way noise impinges on prices differ depending on the area, where in some there is a positive correlation and in others a negative one. Brandt~\cite{brandt_2011} investigates the same hypothesis in the city of Hamburg, Germany by combining multiple sources such as road, air and rail traffic noise pollution with hedonic models too. They highlight the non-linear relationship among noise and price by stating that price decreases significantly lower in areas with low levels of noise, as opposed to high noise level areas where the decrease is more remarkable. Contrary to Brandt’s work, Szczepanska~\cite{szczepanska_2015} study the noise effect on two rather dissimilar locations, with reference to noise, in the city of Olsztyn, Poland. They indicate the existence of linear correlation between prices and noise pollution which underlines the notion that location can influence the noise-price connection in great measure.

Tsao and Lu ~\cite{tsao_2022} collect property data from the Ministry of the Interior of Taiwan for the city of Taoyuan and enhances them with a five year period of noise pollution data from the international airport of Taoyuan. The authors investigate the way aviation noise impacts the real estate market of the city, due to heavy air traffic in lower altitudes, with hedonic models. The models indicate that as the number of flights increases on top of an area, which translates to more noisy conditions, the prices of the corresponding properties decrease noticeably. Moreover, they measure the rate of price decline in certain decibel ranges and conclude that for roughly 65dB of noise due to air traffic the decrease in price can get to 2,356USD, where for more polluted areas the decline reaches the amount of 3,622USD. Similarly, Morano~\cite{morano_2021} study the area of Bari, Italy in order to link noise pollution to house prices, with a total of 200 properties and noise information from the Strategic Noise Map of Bari as well as perceptual views for the quality of an area with regards to noise from residents. To measure the effect of noise, they employ a variation of a data-driven technique known as  Evolutionary Polynomial Regression, or ERP~\cite{bruno_2018}, referred to as ERP-MOGA~\cite{giustolisi_2009} which utilize genetic algorithms. The final results outline the negative correlation between prices and noise levels, where highly polluted areas lead to cheaper housing.

The methodology of the research literature that experimented with the feature of noise in the housing price prediction problem is presented in Table~\ref{tab:noiseLiterature}. Most of them use the conventional hedonic model to predict prices and investigate the contribution of noise. However, there are more advanced techniques available to use in terms of predicting accuracy. Also, there are new practices regarding the interpretability of the model that are able to investigate even further how noise affects the real estate market of a city. These two points are the primary differences of the current work.

\begin{table}[h!]
\caption{Prediction methods used in the literature when incorporating noise}\label{tab:noiseLiterature}
      \begin{tabular}{cc}
        \hline
        \textbf{Paper} & \textbf{Method}  \\ \hline
        Blanco ~\cite{blanco_2011}  & Hedonic Price Model  \\ \hline
        Brandt~\cite{brandt_2011} & Hedonic Price Model  \\ \hline
        Tsao and Lu ~\cite{tsao_2022}  & Hedonic Price Model \\ \hline
        Morano~\cite{morano_2021}  & ERP-MOGA \\ \hline
      \end{tabular}
\end{table}

\section*{Noise Data Reconstruction}
As previously stated, noise data are difficult to obtain because they require specialized equipment for precise measurements, as well as urban environmental specialists capable of completing a task of this complexity. These data must include geographical references in a form of a coordinate system, mapping points or blocks on a map to certain noise values in decibels. As far as we know, there is no such data openly available for the urban area of Thessaloniki, Greece. However, there are official studies of noise pollution for Thessaloniki orchestrated by the Hellenic Ministry of Environment and Energy\footnote{\url{https://ypen.gov.gr/perivallon/thoryvos-aktinovolies/chartografisi-thoryvou-poleodomikon-sygkrotimaton/}}. The studies were conducted in 2015 for three major municipalities of the urban area of Thessaloniki, namely Thessaloniki, Neapoli and Kalamaria, with specialized equipment capable of measuring ground noise caused by factors like vehicles and crowds, while additionally calculating aviation noise produced by airplanes landing to or taking off at the nearby airport. The duration of the studies were set to 46 consecutive days, capturing noise pollution at least once every hour or, in cases, every 15 minutes. The final results were illustrated on a heatmap, where discrete colors represent different noise ranges of 5 decibel intervals. For each municipality, the noise is segmented into daytime and nighttime noise and, in both cases, the data accumulate the overall noise by taking into account both traffic and aviation disturbances. Additionally, for Kalamaria there is a separate heatmap representing only the aviation noise.

\subsection*{Idea and Approach}
The aforementioned studies did not make public the underlying data that were used to create the provided heatmaps. To overcome this problem, we propose a process for reconstructing the initial noise data with a small error. It is important to state that heatmaps used discrete colors mapped to specific small ranges of decibels as shown in Tables~\ref{tab:decibelRangesThessalonikiNeapoli} and \ref{tab:decibelRangesKalamaria}. This means that each color represents the entire range without changing its tone. The ultimate goal is to be able to create the exact same maps by utilizing the reconstructed data. More specifically, the new dataset will contain the noise, in decibels, of a point given its latitude and longitude coordinates.

\begin{table}[h!]
    \caption{Mapping of Noise Ranges to Colors}
    \label{tab:noiseRangesToColorRanges}
    \begin{subtable}[h]{0.45\textwidth}
    \caption{Thessaloniki \& Neapoli Ranges}\label{tab:decibelRangesThessalonikiNeapoli}
    \centering
        \begin{tabular}{cc}
            \hline
            \textbf{Range} & \textbf{Color (RGB)} \\ \hline
            [40, 45) dB & [182, 254, 191] \\ \hline
            [45, 50) dB & [255, 255, 0] \\ \hline
            [50, 55) dB & [254, 196, 71] \\ \hline
            [55, 60) dB & [253, 103, 2] \\ \hline
            [60, 65) dB & [255, 51, 50] \\ \hline
            [65, 70) dB & [152, 0, 51] \\ \hline
            [70, 75) dB & [174, 155, 219] \\ \hline
            [75, 80) dB & [1, 0, 251] \\ \hline
            80+ dB & [1, 1, 65] \\ \hline
          \end{tabular}
    \end{subtable}
    \hfill
    \begin{subtable}[h]{0.45\textwidth}
    \caption{Kalamaria Ranges}\label{tab:decibelRangesKalamaria}
    \centering
        \begin{tabular}{cc}
            \hline
            \textbf{Range} & \textbf{Color (RGB)} \\ \hline
            [35, 40) dB & [80, 167, 50] \\ \hline
            [40, 45) dB & [14, 113, 49] \\ \hline
            [45, 50) dB & [255, 243, 59] \\ \hline
            [50, 55) dB & [172, 121, 78] \\ \hline
            [55, 60) dB & [255, 94, 55] \\ \hline
            [60, 65) dB & [192, 23, 18] \\ \hline
            [65, 70) dB & [138, 18, 19] \\ \hline
            [70, 75) dB & [144, 14, 102] \\ \hline
            [75, 80) dB & [40, 115, 183] \\ \hline
            80+ dB & [10, 65, 121] \\ \hline
      \end{tabular}
    \end{subtable}
\end{table}

It is obvious that an approximation of the noise levels can be extracted through the maps' colors. However, there is no spatial information in order to map each pixel to the corresponding place on a geographic map. To resolve this, we use a technique called Georeferencing~\cite{yao_2020}, which attempts to relate digital representations of maps, like the ones in the noise studies, to a ground system of geographic coordinates. Geographic Information System (GIS) software tools are used to perform this task. In our case, we use an open-source GIS software called QGIS\footnote{\url{https://www.qgis.org/en/site/}}, where by selecting a certain amount of points on the actual map and the corresponding points on the heatmap, we manage to align the two and, eventually, enhance the heatmap with spatial characteristics.

\begin{figure}[h!]
  \caption{\csentence{Color transition effects at the border of decibel ranges (60, 65] and (65, 70] in Thessaloniki's original noise heatmaps}}\label{fig:color_transition}
      \includegraphics[height=2in]{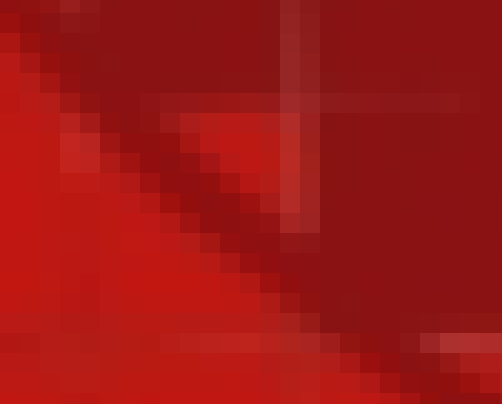}
\end{figure}

The next part of the reconstruction process maps every pixel of the image to a noise value in decibels based on its color. Even though the provided heatmaps incorporate only discrete colors with no fading effects, the pixel-by-pixel analysis introduces resolution restrictions, where many pixels cannot be mapped to a color due to transitioning effects. These cases are found for the most part at the borders where one color switches to another, as depicted in Figure~\ref{fig:color_transition}. To address this problem we can compute the difference between the color of such pixels and the predefined color ranges~\cite{sharma_2005}. When this difference is sufficiently small, we can assign noise values of the corresponding color. If this calculation is done properly, the reconstruction error will be small and, therefore, the data will be more accurate.

One way of proceeding with the calculation is by using the Euclidean distance of the two RGB values. The RGB format consists of 3 separate values ranging from 0 to 255 indicating the amount of red, green and blue in a pixel. State-of-the-art solutions discourage this approach because it does not take into consideration how different these colors are perceived by the human eye~\cite{mokrzycki_2009}. The transition effects on an image aim to smooth the leap from one color to another in terms of human eye perception, which highlights the need of another metric. To overcome this issue, the International Commission on Illumination (CIE) proposed a new method, called $\Delta E^*$, able to handle the perceptual non-uniformities of colors, using the LAB format to represent colors. Similarly to RGB, the LAB format makes use of 3 separate numbers to identify colors. The first one is the Lightness, ranging from 0 to 100, with value 0 defining black and value 100 defining white. The second is the “A” component representing greenness to redness scale, ranging from -128 to +127. Finally, the third is the “B” component which represents blueness to yellowness following the range of the previous component. 

We used the most recent version of the $\Delta E^*$ method, called CIELAB2000. The color comparison result is a number, where 0 means a complete match and as the number increases, the difference between the colors increases too. To use this method, one must carefully select the threshold after which the two colors will be considered to be non-matching. After running some experiments, we set the threshold to 20. As a consequence of the discreteness of colors on the heatmaps, the threshold is not considered to be that crucial in our scenario, because the main goal is to differentiate between the predefined ranges. Essentially, when the color of a pixel matches a color range, this pixel is assigned the corresponding noise value. Since our intention is to correlate housing prices with human perceived noise, we choose to represent each noise range with its arithmetic mean. So, if a pixel color matches the color of the 50-55 range, it will receive 52.5 as its noise value. The final result will be sufficient to describe the noise perception of an area if we take into account the “3dB rule” in the field of Acoustics. The rule states that during an increase of 3 decibels, the sound energy is doubled and, thus, it is accepted as the smallest difference that can be easily heard by most people~\cite{long_2014}. For instance, the average human will rarely notice a transition from 50 to 51 decibels or between 60 and 61.

\subsection*{Process}
The reconstruction process consists of two phases. In the first phase, we scan the image pixel-by-pixel and assign to each pixel the appropriate coordinates, as well as its color. During the second phase, we calculate the color difference of each pixel with the predefined color ranges. It is clear that not all pixels are important due to the transitioning effects we mentioned earlier. For example, in cases where two ranges of radically different colors are adjacent on the map, the transitioning effect will add some pixels in between that probably will not match any color range. Additionally, there are cases where the initial studies could not accurately receive measurements, like the inside of buildings and at the sea. These pixels are not matched to any of the available noise ranges and, hence, are dropped to declutter the data. Table~\ref{tab:finalDatasetStructure} gives the structure of the final dataset.

\begin{table}[h!]
\caption{Final dataset structure}\label{tab:finalDatasetStructure}
      \begin{tabular}{ccc}
        \hline
        \textbf{Features} & \textbf{Type} & \textbf{Range} \\ \hline
        latitude  & float  & [-90, 90]  \\ \hline
        longitude & float  & [-180, 180]  \\ \hline
        red  & int & [0, 255] \\ \hline
        green  & int & [0, 255] \\ \hline
        blue  & int & [0, 255]  \\ \hline
        noise  & float & [0, 85]  \\ \hline
      \end{tabular}
\end{table}

These datasets can be used to create heatmaps that resemble the initial ones. Even though most parts of the images were removed in the process, the remaining locations are still great in number. This can be verified by considering the dataset size in terms of number of rows in the second column of Table~\ref{tab:datasetShapes}. To plot that many points on a single map is exceedingly difficult due to memory constraints. At the same time, the datasets hold spatial information that is way too dense, making them really hard to work with. The dataset supports an accuracy in latitude and longitude values to at least 5 decimals which in actual distances translates to 1.1 meters. This level of detail is unnecessary and superfluous for the purposes of this work. To minimize the density of information to more practical levels, we utilize \textit{tessellation}. Through this method, the map is segmented into separate same-sized squared tiles. We chose to tessellate the map by keeping only the four decimal points of the coordinates. Thus, the accuracy decreases to a resolution of 10 meters that is more manageable and adequate for our case. We group the points based on this rule and aggregate their noise using the arithmetic mean to create a representative indicator for the noise level of the given tile. This technique alters the shape of the dataset as shown in the third column of Table~\ref{tab:datasetShapes} and allows us to plot the results on a map.

\begin{table}[h!]
\caption{Dataset shapes}\label{tab:datasetShapes}
      \begin{tabular}{cccc}
        \hline
        \textbf{Dataset} & \textbf{\# Rows} & \textbf{\# Rows (tessellated)} & \textbf{Reduction} \\ \hline
        Thessaloniki \& Neapoli (Day)  & 3,312,310 & 197,445 & 94\% \\ \hline
        Thessaloniki \& Neapoli (Night) & 3,157,730 & 189,046 & 94\% \\ \hline
        Kalamaria (Day)  & 21,606,947 & 109,245 & 99.4\% \\ \hline
        Kalamaria (Night)  & 20,355,609 & 104,070 & 99.4\%\\ \hline
        Kalamaria Aviation (Day)  & 21,843,537 & 110,111 & 99.4\% \\ \hline
        Kalamaria Aviation (Night)  & 21,831,157 & 109,736 & 99.4\% \\ \hline
      \end{tabular}
\end{table}

\subsubsection*{Results for Thessaloniki and Neapoli}
\begin{figure}[h!]
\caption{\csentence{Average Daily Noise in Thessaloniki/Neapoli}}\label{fig:thessaloniki_day}
\centering
\begin{subfigure}{.5\textwidth}
  \centering
  \includegraphics[width=1\linewidth]{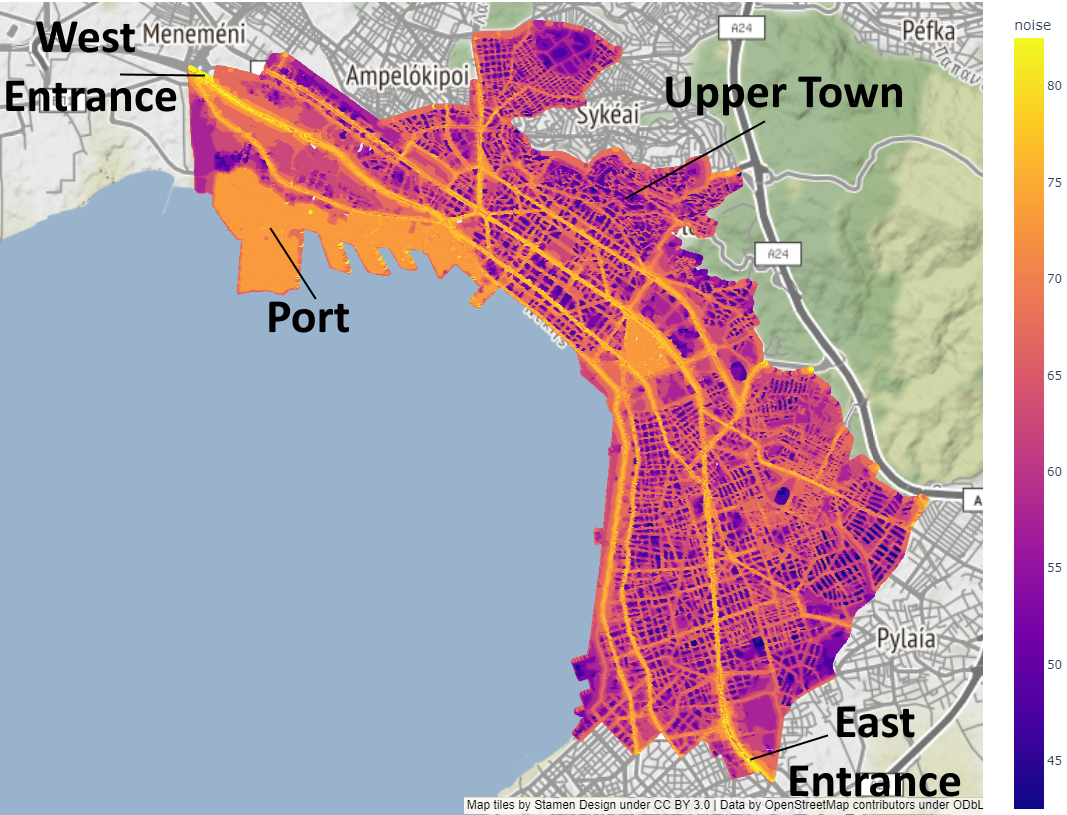}
  \caption{Reconstructed}
  \label{fig:thessaloniki_day_new}
\end{subfigure}%
\begin{subfigure}{.5\textwidth}
  \centering
  \includegraphics[width=1\linewidth]{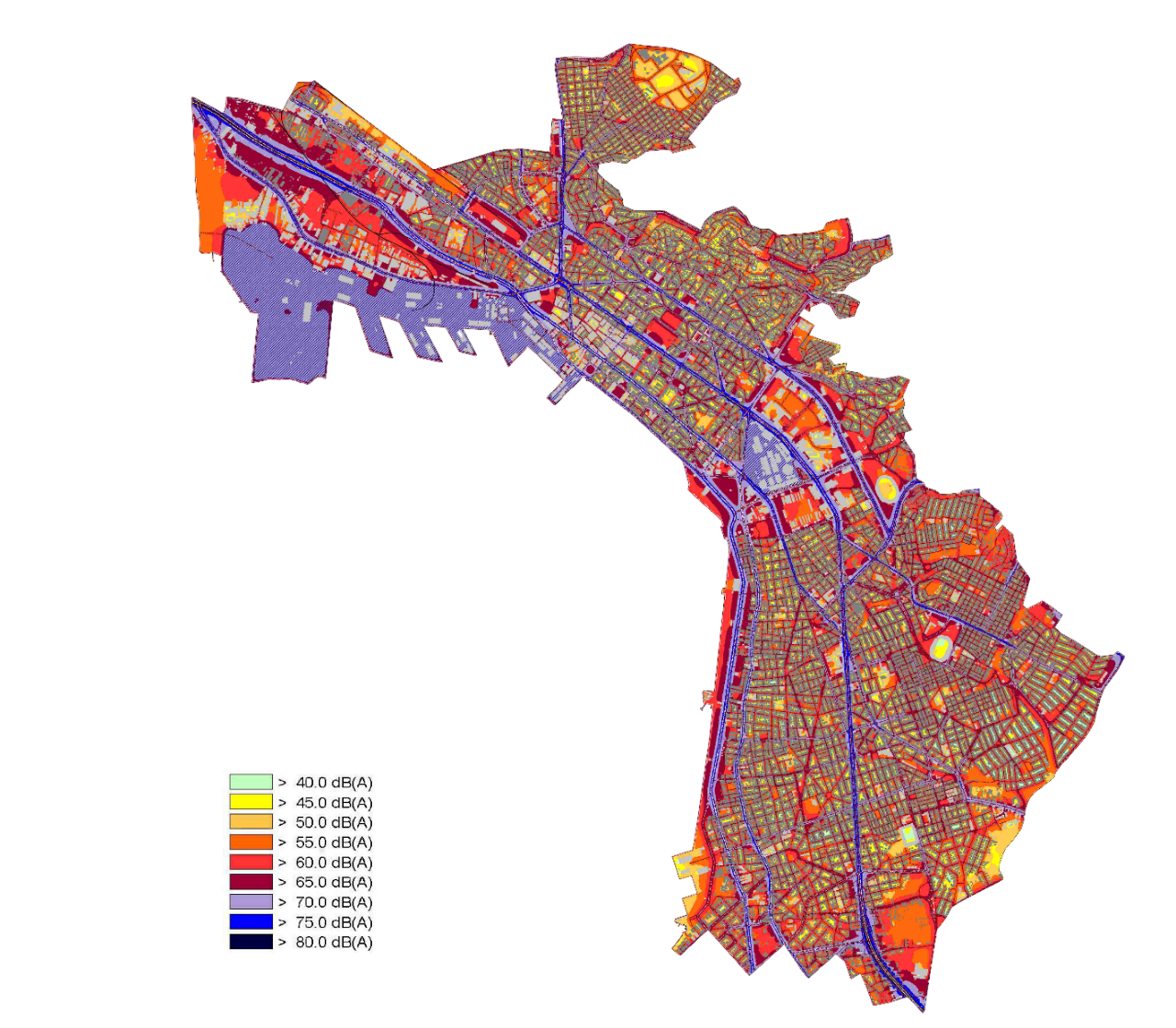}
  \caption{Original}
  \label{fig:thessaloniki_day_original}
\end{subfigure}
\end{figure}

\begin{figure}[h!]
\caption{\csentence{Average Nightly  Noise in Thessaloniki/Neapoli}}\label{fig:thessaloniki_night}
\centering
\begin{subfigure}{.5\textwidth}
  \centering
  \includegraphics[width=1\linewidth]{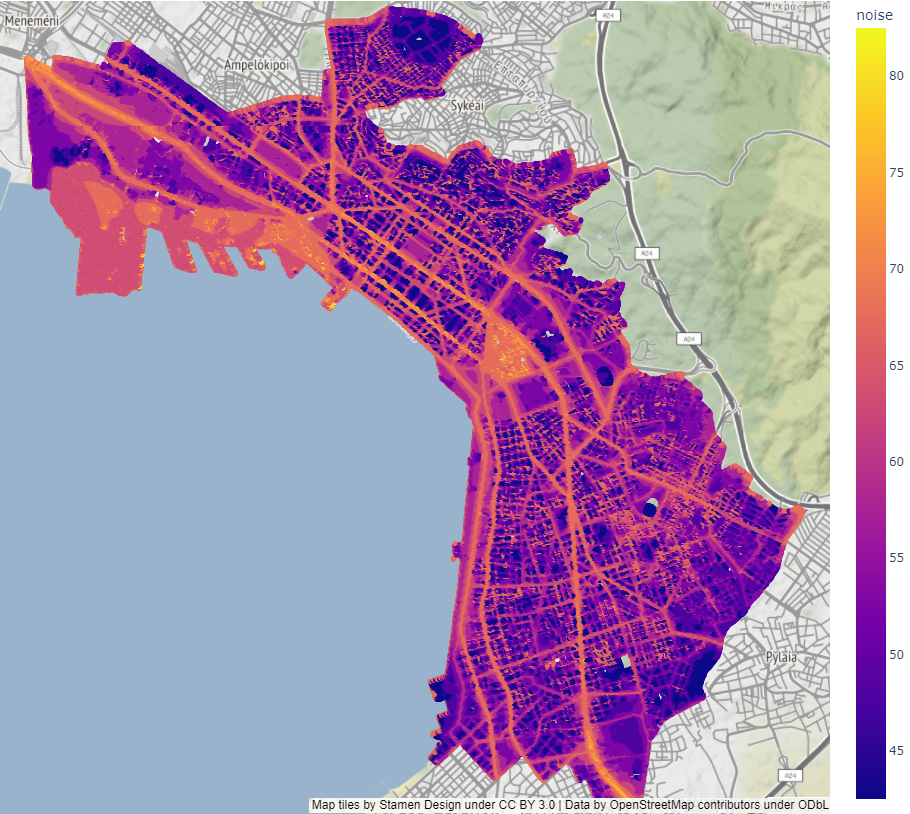}
  \caption{Reconstructed}
  \label{fig:thessaloniki_night_new}
\end{subfigure}%
\begin{subfigure}{.5\textwidth}
  \centering
  \includegraphics[width=1\linewidth]{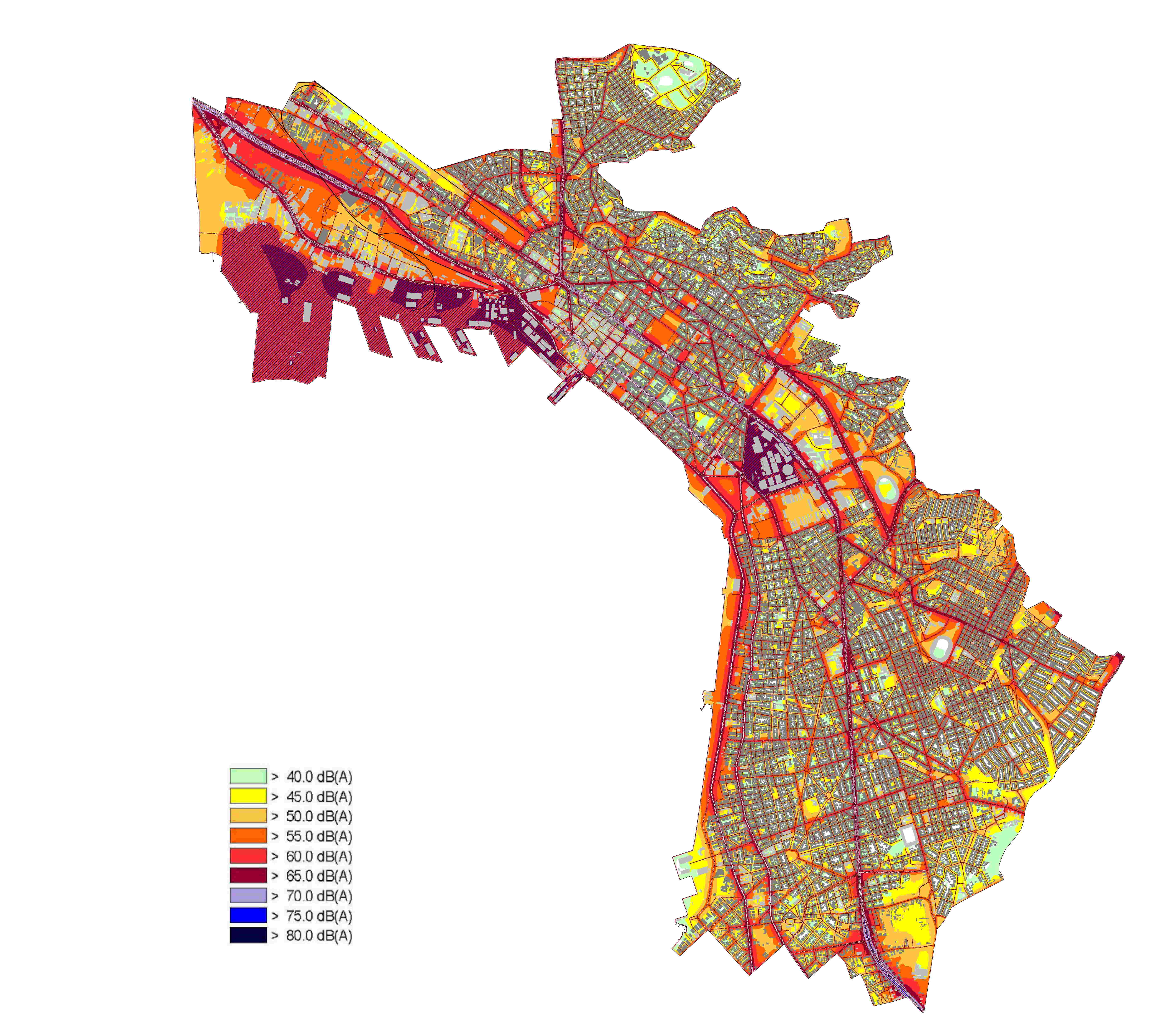}
  \caption{Original}
  \label{fig:thessaloniki_night_original}
\end{subfigure}
\end{figure}

Figure~\ref{fig:thessaloniki_day} shows the average daily noise for the areas of Thessaloniki and Neapoli, ranging from 40dB to almost 85dB, for both the original and the reconstructed versions. The noisiest parts are the main roads and the intersections that can accommodate large numbers of vehicles. The two most distinguishable examples are the East and West entrances of the city where the noise can reach a level of 80dB. Also, it is visible the way that the noise spreads almost equally around these highly polluted spots, which, in fact, increase the noise pollution of the surrounding area. Besides the road network, one more part of the city that is apparently noisy is the port, which is very big in size and greatly active during both daytime and nighttime. Furthermore, the correlation between road size, which leads to high traffic, and the noise pollution can be validated in urban areas with narrow streets. A very useful example is the area of “Upper Town” marked in Figure~\ref{fig:thessaloniki_day}, which is one of the oldest parts of the city where due to the increased elevation and the rough terrain the roads are extremely narrow. This fact, except the restrictions it imposes on the number of vehicles that can pass simultaneously, makes access difficult and not appealing to drivers. This is one reason why it is one of the quietest places in Thessaloniki. Figure~\ref{fig:thessaloniki_night} shows the average nightly noise in the same area in which, although the noisiest and quietest places remain the same, the noise pollution levels are much lower.

\subsubsection*{Results for Kalamaria}
As in the previous subsection, Figure~\ref{fig:kalamaria_day} shows the average daily noise for the area of Kalamaria. Once again, the noisiest places are the main roads, while the quietest are those surrounded by low traffic streets. The yellow color indicates regions with the maximum noise levels such as the core intersections. Contrary to the heatmaps of the other two municipalities where the noise was almost entirely driven by the road network, in Kalamaria there are certain zones with little or no road network that are very noisy. This is caused due to air traffic, since the airline routes pass over the vicinity in relatively low altitude and the turbines generate noise that can reach over 100dB~\cite{Konopka_2014}. This effect is more recognizable at night (see Figure~\ref{fig:kalamaria_night}). Despite the fact that the road network has minimal traffic, some areas are noisier compared to others. The noise generated by airplanes is shown in Figure~\ref{fig:kalamaria_aviation_day} and~\ref{fig:kalamaria_aviation_night}. These figures are zoomed in a bit to improve readability and distinguish the street layout. The aviation data can be of great interest both in research and in industry, so in this paper, we provide a separate dataset for the aviation noise.

\section*{Implementation and Experimentation}
\subsection*{Property Data}
Investigating the correlation and influence of noise in housing prices requires a real world housing prices dataset. For the purposes of this paper, we have utilized  Openhouse\footnote{\url{https://openhouse.gr}}. Openhouse is a real estate platform operating in major cities of Greece. It contains high quality information for a wide range of properties, considering multiple aspects of them. Since Openhouse is a data oriented platform, paying critical attention to their service, they have provided Thessaloniki’s properties in order to experiment with the noise data reconstructed in the previous section. The data refer to residential properties offered for sale that were listed on the platform in October 2022. Each property has the features mentioned in Table~\ref{tab:propertyFeatures}.

\begin{table}[h!]
\caption{Openhouse Property Features}\label{tab:propertyFeatures}
      \begin{tabular}{cccc}
        \hline
        \textbf{Feature} & \textbf{Type} & \textbf{Missing Values} & \textbf{Imputation} \\ \hline
        Size & Float & 0\% & -  \\ \hline
        NumberOfRooms & Int & 0\% & -  \\ \hline
        Latitude & Float & 0\% & -  \\ \hline
        Longitude & Float & 0\% & -  \\ \hline
        EnergyEfficiencyId & Categorical  & 0\% & -  \\ \hline
        ConstructionDate & Datetime & 13.75\% & Mean  \\ \hline
        SubTypeId & Categorical & 0.17\% & Mode  \\ \hline
        FloorLevelId & Categorical & 0.30\% & Rounded mean  \\ \hline
        BasicHeatingTypeId & Categorical & 30.71\% & Mode  \\ \hline
        DoorFrameTypeId & Categorical & 31.77\% & Mode  \\ \hline
      \end{tabular}
\end{table}

The majority of the features are self-explanatory with the exception of `SubTypeId' and `DoorFrameTypeId'. `SubTypeId' refers to the structural subtype of the residential property receiving values like `apartment' and `studio' among others. `DoorFrameTypeId' corresponds to the type of door frames a property has, such as `synthetics' and `aluminum' to name a few. We have performed an exploratory data analysis on the given dataset to locate potential outliers and verify the overall integrity. Outlier detection was done with the interquartile range (IQR) method. Using IQR in the `NumberOfRooms' feature led to an upper limit of 7 rooms, which decreased the dataset size by no more than 1\%. Similarly, in the `Size' feature, the upper limit was 300$m^2$, which consequently reduced the size by almost 8\%. Additionally, price outliers were removed too, by forcing a price range between 10,000 and 500,000 euros. Eventually, the filtered set consists of 2,014 properties. The missing values were filled according to Table~\ref{tab:propertyFeatures}, where different aggregations were used depending on the data type. It must be noted that although `DoorFrameTypeId' and `BasicHeatingTypeId' features are missing approximately 30\% of their values, they are considered of significant importance in the housing price prediction process based on the domain knowledge provided by Openhouse. Therefore, we decided to fill these too, and check their influence in practice. As far as the encoding of features, the `EnergyEfficiencyId' and `FloorLevelId' were encoded using incremental indices because they are ordinal categorical features. The other categorical features are nominal so one-hot and binary encoding~\cite{potdar_2017} were used and compared. The one-hot encoding achieved better results and, thus, used in the following experiments.

\subsection*{Experiments}
To investigate the correlation between housing prices and noise we utilize tree-based models that perform well in similar cases ~\cite{truong_2019,kang_2020,zou_2022}. In particular, we use decision trees, random forest, XGBoost and light gradient boosting models. To verify the impact of noise we employ standard interpretability methods like feature importance, partial dependence~\cite{hastie_2008, greenwell_2018} and permutation importance~\cite{altmann_2010} plots. To shed even more light on interpretability, we employ other advanced techniques such as local interpretable model-agnostic explanations, or LIME,~\cite{ribeiro_2016} and Shapley additive explanations, or SHAP,~\cite{lundberg_2017}. The hyperparameter tuning for each model was accomplished with Bayesian optimization~\cite{snoek_2012}, which outperformed grid search, and 5-fold cross-validation. 

The experiments were structured in three different axes. The first one corresponds to the procedure followed to assign the appropriate noise value to each property of the dataset. We choose to average the noise within a certain radius around each property, where the actual radius distance is manually selected. The second one refers to the main noise characteristics we can use when assigning a noise value to a property. These characteristics are the following:

\begin{itemize}
    \item One feature for the average day noise and one for the average night noise (I)
    \item One feature which averages both day and night noise (II)
    \item One feature for the average day noise (III)
    \item One feature for the average night noise (IV)
\end{itemize}

\begin{figure}[h!]
  \caption{\csentence{Areas of Thessaloniki}}\label{fig:areas}
      \includegraphics[height=2.3in]{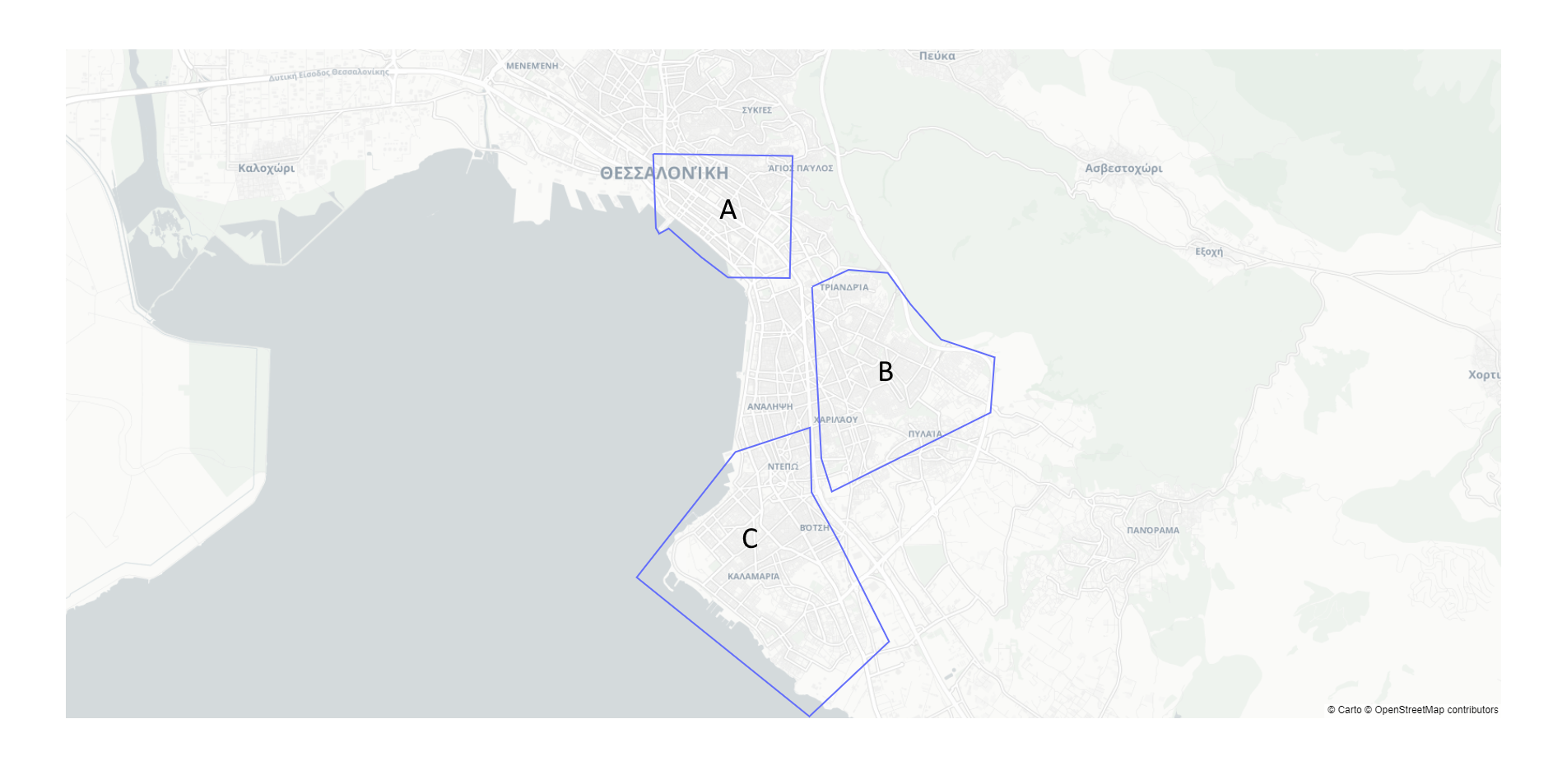}
\end{figure}

The third and last experimental component is the area where we examine the effect of noise in pricing. The presence of noise can be translated differently depending on the urban attributes of each part of a city. Good examples that demonstrate this behavior are city centers, where the noise levels are usually increased compared to other places in the same city as a consequence of the high road and pedestrian traffic. In turn, the traffic is caused by the commercial nature of the center since most of the provided services and amenities are located there. This, in fact, gives extra value to the properties of this area which makes them more expensive. However, this is not the case in other parts of the city. For instance, in the suburbs, where there are mostly residential properties of families, the absence of noise is generally considered to be a positive factor that can raise the prices. Taking these into consideration, we focus on three different areas of Thessaloniki with contrasting urban features: the city center (A), Triandria, Toumpa and Harilaou areas (B) and Kalamaria area (C), as they are depicted in Figure~\ref{fig:areas}. Another reason why we chose these areas is their difference in terms of price-noise correlation. This is illustrated in Figure~\ref{fig:corr_noise_price}, where the correlation between price per $m^2$ and noise is plotted for the entire area of interest as well as each individual area. Even though it is difficult to imply any kind of correlation when looking at the entire area of Thessaloniki, that is not the case for areas A and C, where their shapes indicate some sort of correlation. Area B follows the pattern of Figure~\ref{fig:corr_all} and was chosen as a representative subset of the entire area.

\section*{Results}
The final results are organized into two different groups based on the noise radius that was used. These radiuses were set to 100 and 50 meters. For each group all four models were trained on the three areas of Thessaloniki for all four noise characteristics described in the previous section. Due to the area segmentation, the number of properties has declined, leading to a concern about the sufficiency of the training set. To make sure the data were enough to be able to make valid conclusions, we plotted the learning curves of each model and verified that the curves reach a plateau. Also, because of the large number of different experimental combinations based on the experimentation axes we mentioned previously, we decided to omit showcasing every examination of the noise characteristics and keep, only, the one that performs the best. We should point out that when changing noise radius there are circumstances where a property can end up without a noise value, especially when the radius decreases. In such cases, these properties are removed from the dataset and this is why there seems to be inconsistencies in the results when switching from one radius to another, even without incorporating the noise data.

\begin{table}[h!]
\caption{Results for radius set to 100m}\label{tab:resultsR100}
      \begin{tabular}{|c|ccc|ccc|ccc|}
        \hline
         & \multicolumn{3}{c}{\textbf{A}} & \multicolumn{3}{|c}{\textbf{B}} & \multicolumn{3}{|c|}{\textbf{C}} \\ \hline
         \textbf{Model} & \textbf{MAE} & \textbf{MAPE} & \textbf{Noise} & \textbf{MAE} & \textbf{MAPE} & \textbf{Noise} & \textbf{MAE} & \textbf{MAPE} & \textbf{Noise}\\ \hline
        XGBoost & 28919 & \textbf{0.223} & - & 22504 & 0.15 & - & 31511 & 0.138 & -  \\ \hline
        XGBoost $\dagger$ & \textbf{28888} & 0.233 & II & \textbf{19189} & \textbf{0.144} & I & \textbf{30141} & \textbf{0.128} & IV  \\ \hline
        LGBM & 32572 & 0.258 & - & 21618 & 0.158 & - & 32752 & 0.151 & -  \\ \hline
        LGBM $\dagger$ & 31477 & 0.258 & II & 22715 & 0.173 & II & 31139 & 0.138 & IV  \\ \hline
        RF & 32519 & 0.267 & - & 24259 & 0.181 & - & 38922 & 0.165 & -  \\ \hline
        RF $\dagger$ & 31759 & 0.259 & II & 24481 & 0.182 & II & 40389 & 0.173 & II  \\ \hline
        DT & 35264 & 0.277 & - & 28966 & 0.238 & - & 48802 & 0.209 & -  \\ \hline
        DT $\dagger$ & 31771 & 0.271 & III & 30671 & 0.25 & II & 52077 & 0.225 & III  \\ \hline
        \multicolumn{10}{c}{} \\
        \multicolumn{10}{c}{\textsuperscript{$\dagger$}:\footnotesize{Indicates the use of noise.}} \\
        \multicolumn{10}{c}{\footnotesize{\textbf{Bold} text marks the best score across all models for a given area.}} \\
      \end{tabular}
\end{table}

\begin{table}[h!]
\caption{Results for radius set to 50m}\label{tab:resultsR50}
      \begin{tabular}{|c|ccc|ccc|ccc|}
        \hline
         & \multicolumn{3}{c}{\textbf{A}} & \multicolumn{3}{|c}{\textbf{B}} & \multicolumn{3}{|c|}{\textbf{C}} \\ \hline
         \textbf{Model} & \textbf{MAE} & \textbf{MAPE} & \textbf{Noise} & \textbf{MAE} & \textbf{MAPE} & \textbf{Noise} & \textbf{MAE} & \textbf{MAPE} & \textbf{Noise}\\ \hline
        XGBoost & 28919 & \textbf{0.223} & - & 22504 & 0.15 & - & 31511 & 0.138 & -  \\ \hline
        XGBoost $\dagger$ & \textbf{28001} & 0.229 & I & \textbf{20858} & \textbf{0.15} & I & 31370 & 0.132 & III  \\ \hline
        LGBM & 32572 & 0.258 & - & 21618 & 0.158 & - & 32752 & 0.151 & -  \\ \hline
        LGBM $\dagger$ & 30216 & 0.241 & III & 22408 & 0.161 & IV & \textbf{29872} & \textbf{0.13} & III  \\ \hline
        RF & 31785 & 0.256 & - & 24224 & 0.182 & - & 38886 & 0.165 & -  \\ \hline
        RF $\dagger$ & 31380 & 0.254 & II & 24028 & 0.183 & I & 39626 & 0.168 & IV  \\ \hline
        DT & 35319 & 0.279 & - & 33561 & 0.27 & - & 48802 & 0.209 & -  \\ \hline
        DT $\dagger$ & 35453 & 0.271 & III & 27756 & 0.191 & III & 50290 & 0.209 & II  \\ \hline
      \end{tabular}
\end{table}

The results of Table~\ref{tab:resultsR100}, where the radius is set to 100 meters, indicate a clear dominance of the XGBoost model in terms of both mean squared error (MAE) and mean absolute percentage error (MAPE) values. The performance gain in each area varies as well as the noise characteristics that are used. More precisely, in area A there is no significant improvement, while in the other two areas noise improves both scores radically. The LGBM model benefits from noise only in area C. The random forest and decision tree models are unable to make use of noise with the exception of area A where both are boosted. When the radius is set to 50 meters in Table~\ref{tab:resultsR50}, we observe the same pattern where the hierarchy between the models remains the same. The main differences appear to be the LGBM model that achieves finer results than XGBoost in area C and, also, the decision tree which is crucially improved with the use of noise in area B. Regarding the best performing models, even though setting the radius to 50 meters can reduce the MAE in areas A and C, the MAPE does not change remarkably. Furthermore, decreasing the radius exacerbates the results in area B, so the radius switch does not necessarily enhance the overall performance of the model.

To measure the extent by which noise increases model performance and investigate the correlation between noise and price through interpretability evaluation methods, for the best performing models of each area, we plot the feature importance, permutation importance and partial dependence plots together with LIME and SHAP plots. We must mention that in permutation importance plots the measure of importance in XGBoost refers to the average gain across all splits a feature is used in, while in LGBM refers to the number of times a feature is used to split the data.

\subsection*{Area A}
For the center area of Thessaloniki XGBoost is the best performing model, when radius is set to 50 meters. In this model, both average day and night noise are used as features in the training. The average day noise is ranked in the feature importance plot of Figure~\ref{fig:r50A_featureImportance} almost as high as the construction date, while the night noise is located at a couple of ranks below. In the same plot, the `SubTypeId\_4', which denotes properties classsified as studios, is marked as the most important feature. The partial dependence plot in Figure~\ref{fig:r50A_day} shows that property prices increase as the noise increases, which confirms the initial claim that city centers evaluate noise positively, which most probably occurs due to their commerciality. This can be verified by the LIME weights in Figure~\ref{fig:r50A_lime} where high noise values correspond to bigger weights. SHAP values in the beeswarm of Figure~\ref{fig:r50A_shap} highlights this relationship too, since the left hand-side is mostly colored with blue (low values), while the right hand-side with red (high values). At last, in Figure~\ref{fig:r50A_night} the night noise does not appear to act on prices at the quieter areas. However, as we progress to noisier parts, night noise has a negative impact on pricing. This is not strange because during night time the commerciality factor is not that crucial.

\subsection*{Area B}
For the Triandria, Toumpa and Harilaou areas the best results are demonstrated, again, by XGBoost with a noise radius of 100 meters. As in the previous area, this model utilizes both average day and night noise values. The night noise has similar importance to features such as the location and the heating type as it is depicted in Figure~\ref{fig:r100B_featureImportance}. In the same figure, the permutation importance plot showcases that the overall noise affects at some degree the accuracy of the model. Even though, at first, day and night noise do not seem to influence prices, after a certain threshold in decibels they do have a negative effect on prices which contradicts the results of area A. One of the possible reasons why noise does not cause price changes in the initial decibel ranges is the fact that some parts of area B are close to the city center and, hence, noise is not directly considered as a bad attribute. Once more Figures~\ref{fig:r100B_lime} and~\ref{fig:r100B_shap} reinforces the previous findings about the generally negative correlation between noise and price. It should be noted that area B is the only area where setting the radius to 100 meters leads to better results when compared to setting it to 50 meters. This differentiation can be attributed to the fact that area B is the only one located far from the coastline. Seaside properties are notably more expensive than the rest, which makes property prices' distribution less uniform. In such scenarios, using a smaller radius can help the model recognize these non-uniformities. Due to area's B distance from the sea, this does not occur, making the selection of a bigger radius a better choice.

\subsection*{Area C}
In the Kalamaria area, the LGBM model when trained with a noise radius of 50 meters while taking into consideration only the average day noise achieves the best scores. The performance gain in terms of MAE is at approximately 2.880 euros and for MAPE marginally over 2\%. In particular, the noise is ranked almost as high as `Size' with regard to importance in Figure~\ref{fig:r50C_featureImportance}. This area is located far from the center and as a consequence the noise appears to influence price negatively at most noise ranges. In Figure~\ref{fig:r50C_day}, the price declines almost linearly as we move to more noisy parts of the area, while LIME weights in Figure~\ref{fig:r50C_lime_day} indicate the preference of the model to assign higher prices to properties with relatively low surrounding noise. Once more, the SHAP values of Figure~\ref{fig:r50C_shap} confirm the aforementioned observations, where high average day noise values cause price drops and low average noises escalate prices. As far as the noise characteristic used, one plausible reason why the model chooses to incorporate only the day noise is that contrary to the previous areas, Kalamaria includes also the aviation noise. As it can be seen by the corresponding heatmaps, aviation noise during night increases the overall noise which at some extend narrows the gap between day and night noise. This means that the two noise features are more correlated and, thus, one of them can potentially be redundant.

\section*{Discussion and Conclusion}
The main goal of this paper was to investigate how urban noise impacts residential property prices in the area of Thessaloniki. Currently, there is no publicly available spatial data regarding noise for the area of interest. Therefore, the first part of this work attempts to create a general purpose dataset indicating the sense of noise based on coordinates by taking advantage of official and public studies conducted by the Hellenic Ministry of Environment and Energy. 

This new dataset is combined with the properties of the Openhouse platform to train tree-based machine learning models in order to verify the importance of noise in housing price estimates. The assumption that noise might be translated differently depending on the location of the property led us to focus the experiments on three separate regions of Thessaloniki with dissimilar characteristics. XGBoost and LGBM models attain the best results which first of all confirm that noise, as a matter of fact, influences prices and, secondly, it can affect some locations positively while others negatively. More specifically, property prices in the city center as well as locations in its vicinity, do increase as noise increases, which is probably the aftereffect of the overall commerciality of the area. In contrast, properties located far from the center are impacted negatively by noise. This makes sense considering that in decentralized areas, such as suburbs, there are mainly houses of families where quietness is more appreciated. 

Besides the further research that can be conducted around this topic with different models, different property types and features, the newly reconstructed noise dataset can be used wherever needed for commercial projects and researches that are, even, not real estate related, since it is a general purpose sense of noise set.

\clearpage

\appendix
\section{Reconstructed Heatmaps}
\begin{figure}[h!]
  \caption{\csentence{Average Daily Noise in Kalamaria}}\label{fig:kalamaria_day}
      \includegraphics[height=2.1in]{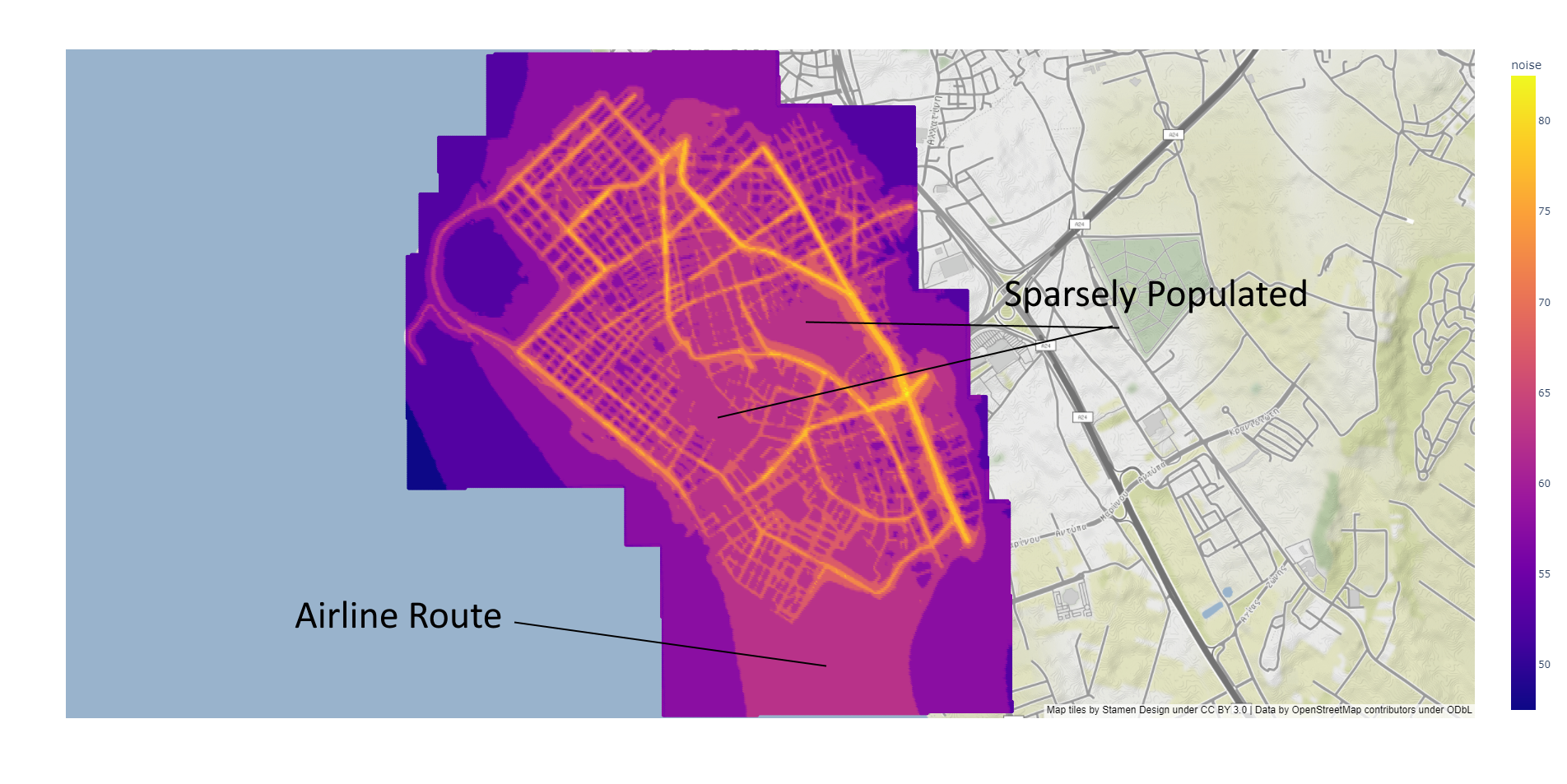}
\end{figure}

\begin{figure}[h!]
  \caption{\csentence{Average Nightly Noise in Kalamaria}}\label{fig:kalamaria_night}
      \includegraphics[height=2.1in]{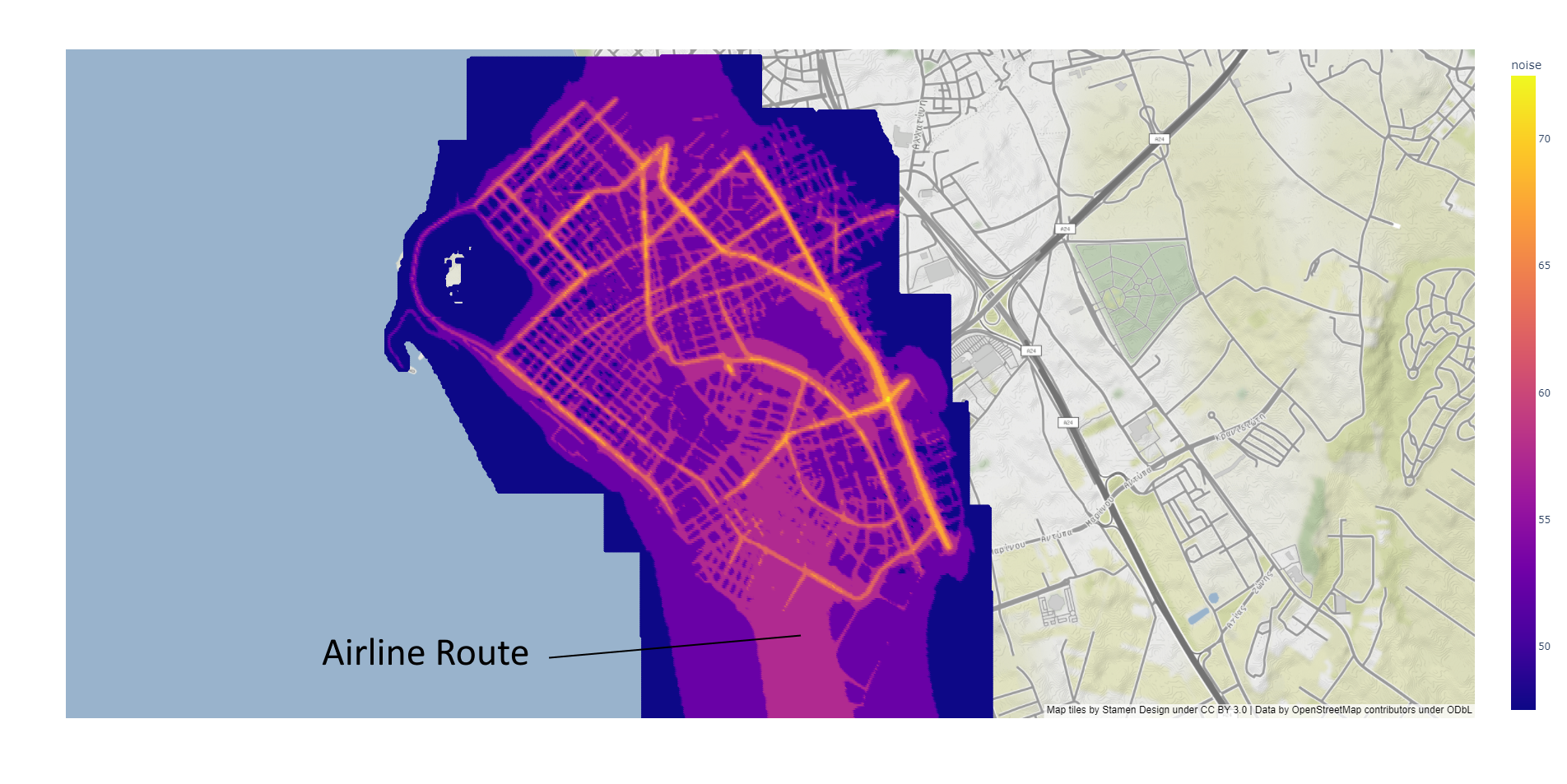}
\end{figure}

\begin{figure}[h!]
  \caption{\csentence{Average Daily Aviation Noise in Kalamaria}}\label{fig:kalamaria_aviation_day}
      \includegraphics[height=2.1in]{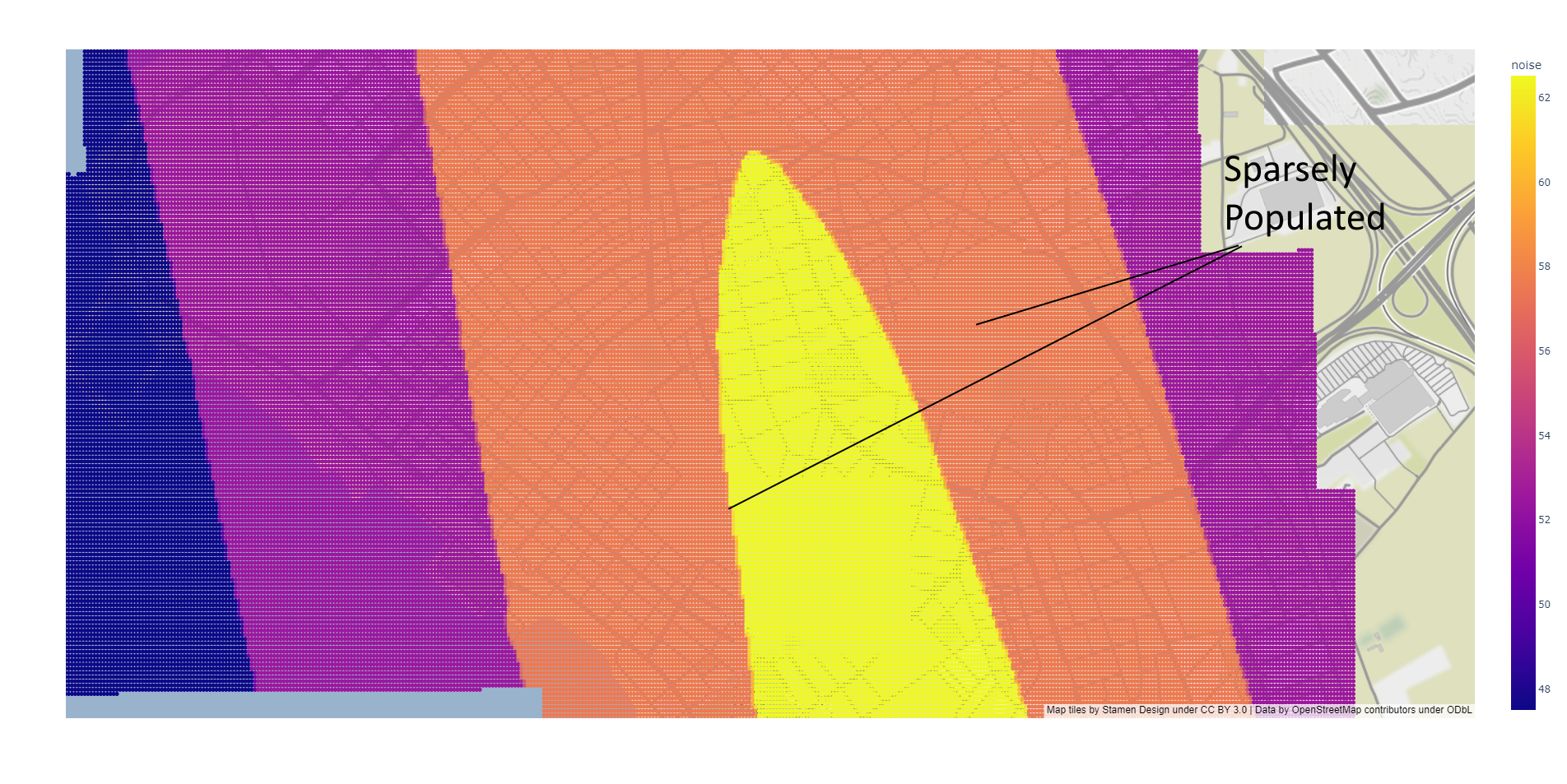}
\end{figure}

\begin{figure}[h!]
  \caption{\csentence{Average Nightly Aviation Noise in Kalamaria}}\label{fig:kalamaria_aviation_night}
      \includegraphics[height=2.1in]{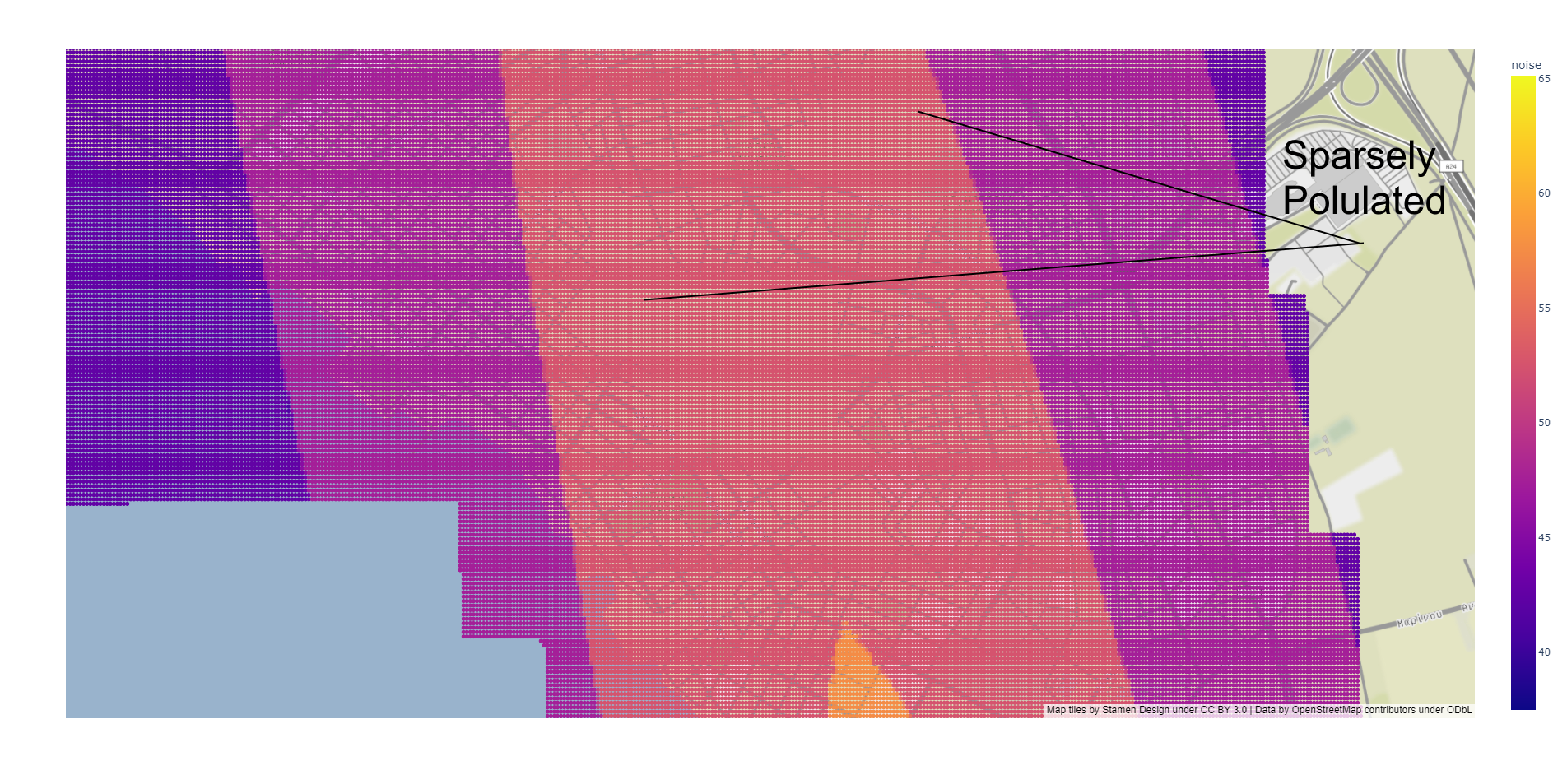}
\end{figure}

\section{Correlation Plots}
\begin{figure}[h!]
\caption{\csentence{Correlation Between Price/m$^2$ and Noise for Different Areas}}\label{fig:corr_noise_price}
\begin{subfigure}{.475\linewidth}
  \includegraphics[width=\linewidth]{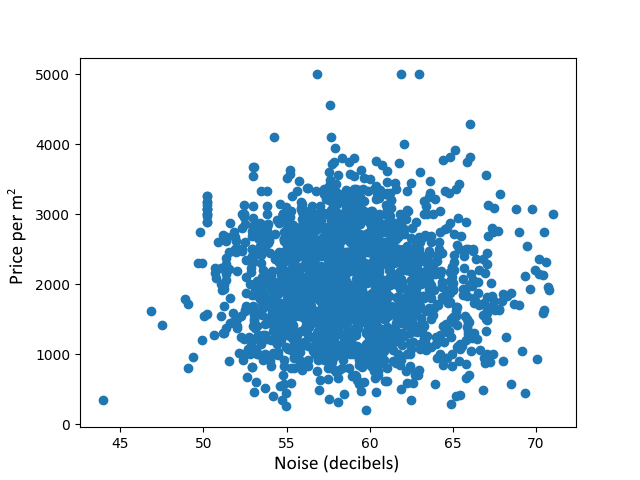}
  \caption{Entire Area of Thessaloniki}
  \label{fig:corr_all}
\end{subfigure}\hfill 
\begin{subfigure}{.475\linewidth}
  \includegraphics[width=\linewidth]{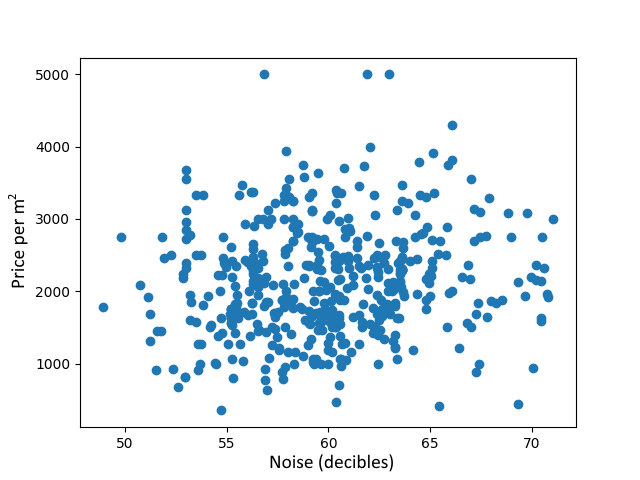}
  \caption{Center (A)}
  \label{fig:corr_A}
\end{subfigure}

\medskip 
\begin{subfigure}{.475\linewidth}
  \includegraphics[width=\linewidth]{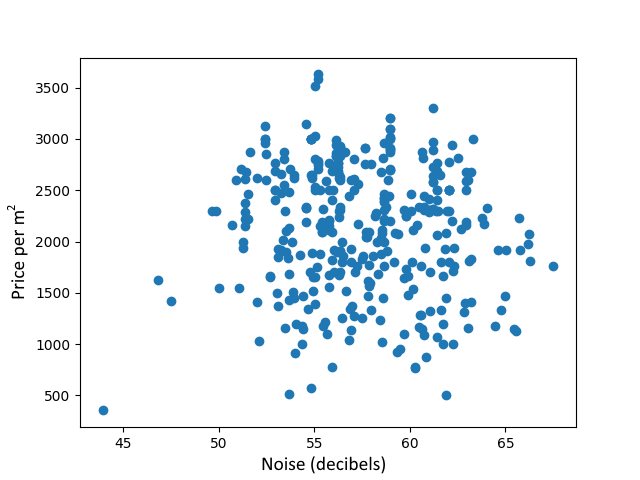}
  \caption{Triandria, Toumpa and Harilaou (B)}
  \label{fig:corr_B}
\end{subfigure}\hfill 
\begin{subfigure}{.475\linewidth}
  \includegraphics[width=\linewidth]{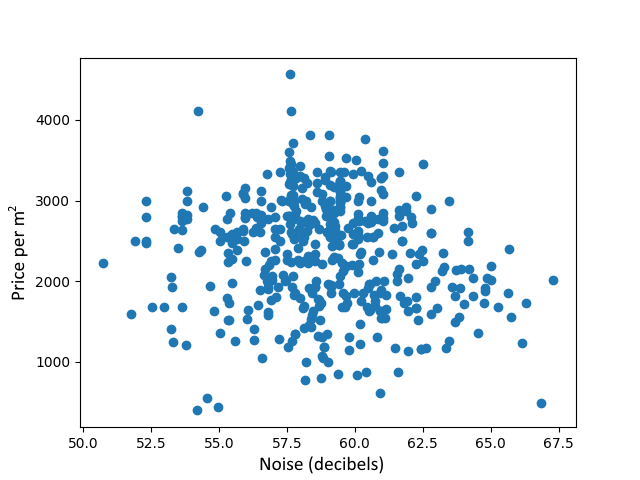}
  \caption{Kalamaria (C)}
  \label{fig:corr_C}
\end{subfigure}
\end{figure}

\clearpage

\section{Result Plots}
\begin{figure}[h!]
  \caption{\csentence{Feature Importance \& Permutation Importance for Area A}}\label{fig:r50A_featureImportance}
      \includegraphics[height=2.1in]{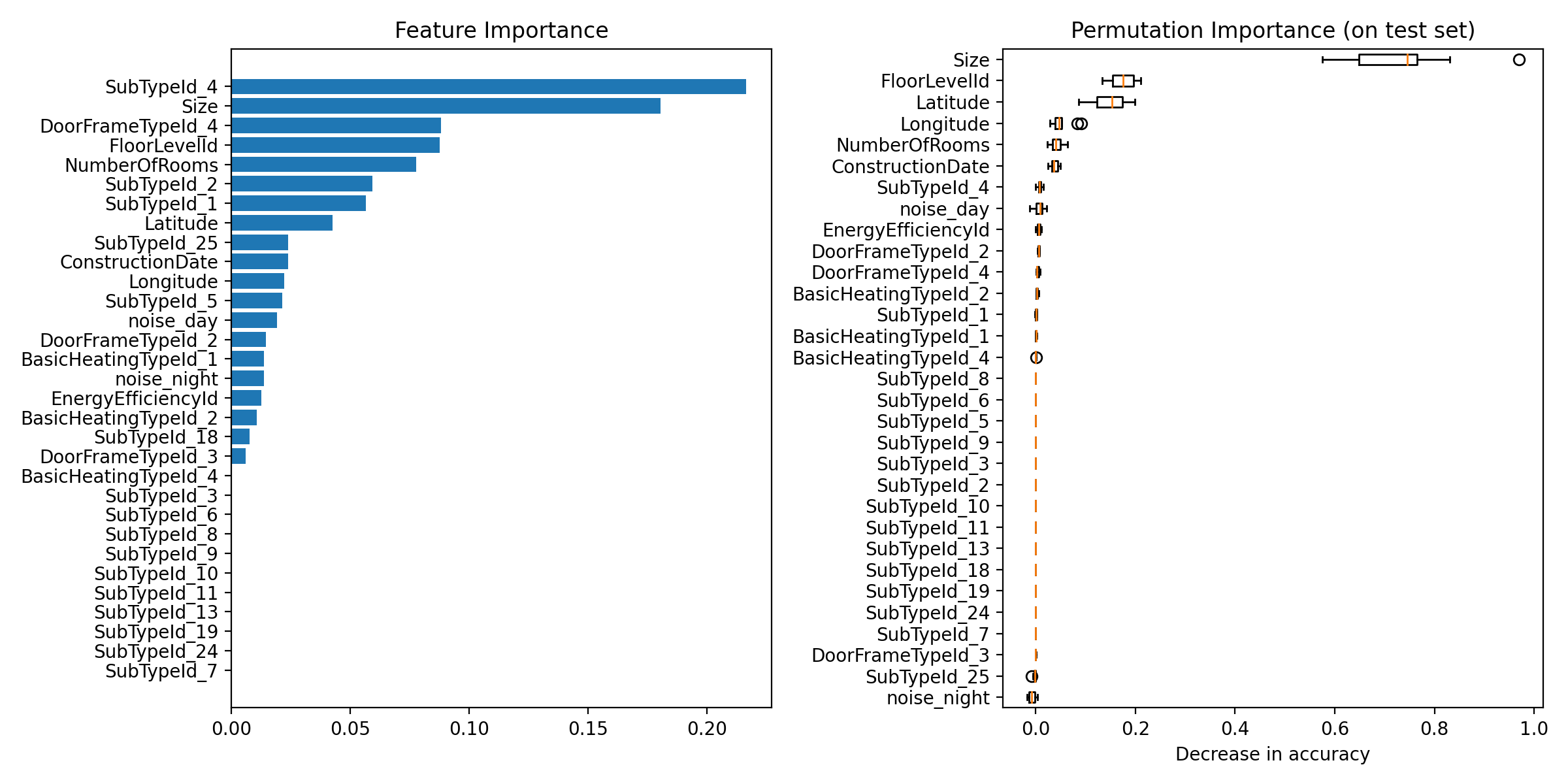}
\end{figure}

\begin{figure}[h!]
\caption{\csentence{Partial Dependence Plots for Area A}}\label{fig:r50A_PDP}
\centering
\begin{subfigure}{.5\textwidth}
  \centering
  \includegraphics[width=1\linewidth]{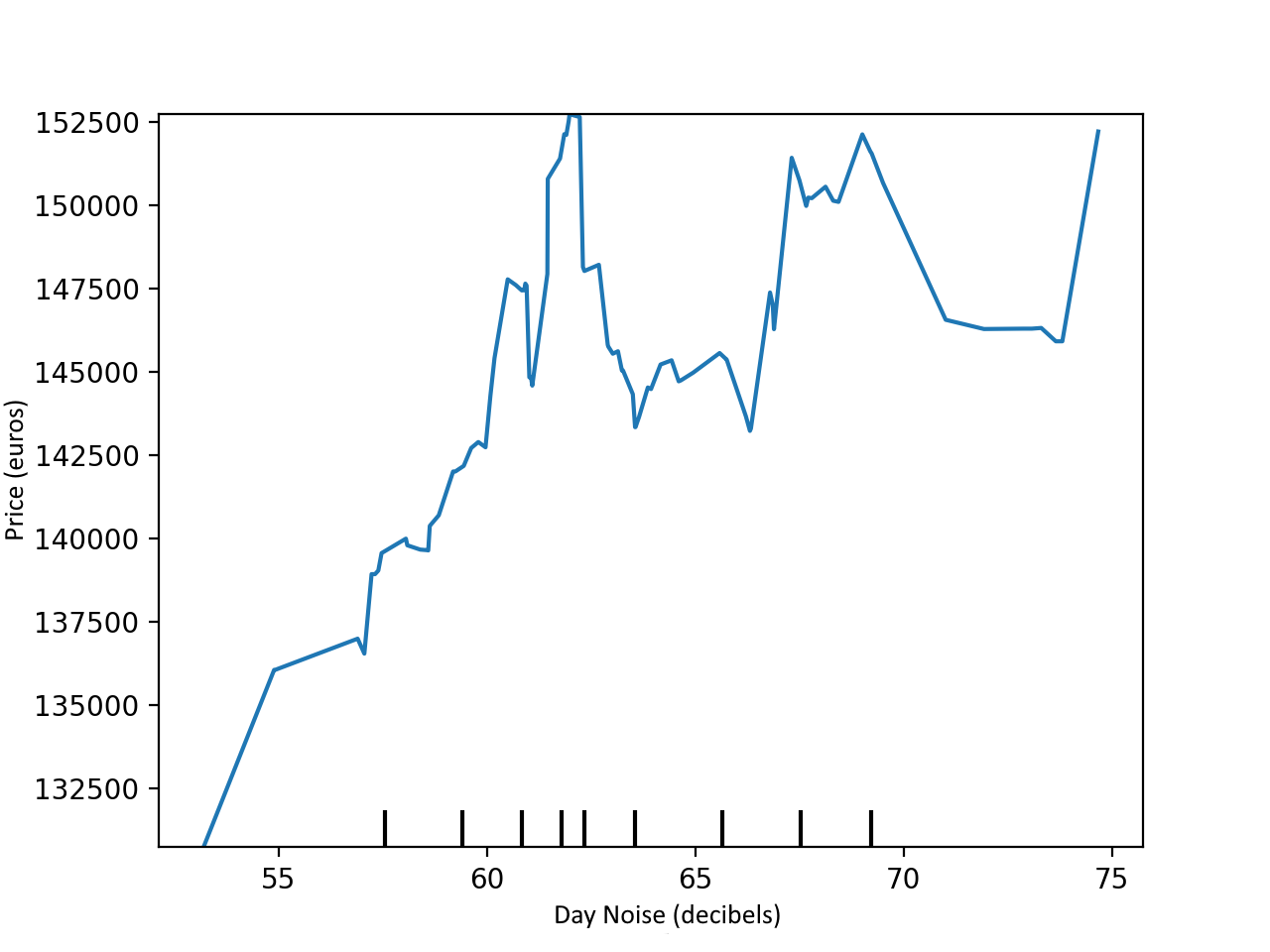}
  \caption{Day Noise}
  \label{fig:r50A_day}
\end{subfigure}%
\begin{subfigure}{.5\textwidth}
  \centering
  \includegraphics[width=1\linewidth]{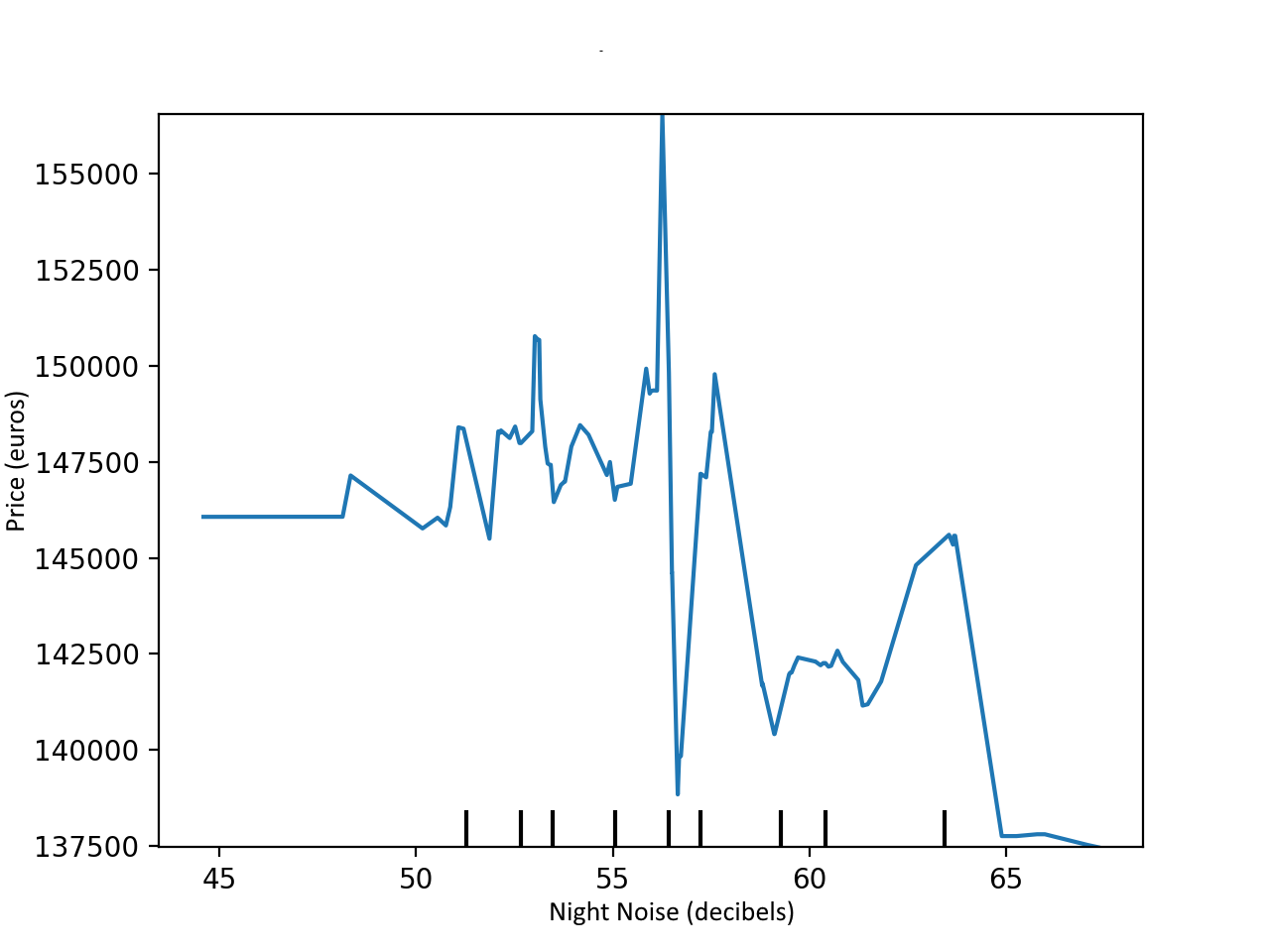}
  \caption{Night Noise}
  \label{fig:r50A_night}
\end{subfigure}
\end{figure}

\begin{figure}[h!]
  \caption{\csentence{SHAP Plot for Area A}}\label{fig:r50A_shap}
      \includegraphics[height=2.1in]{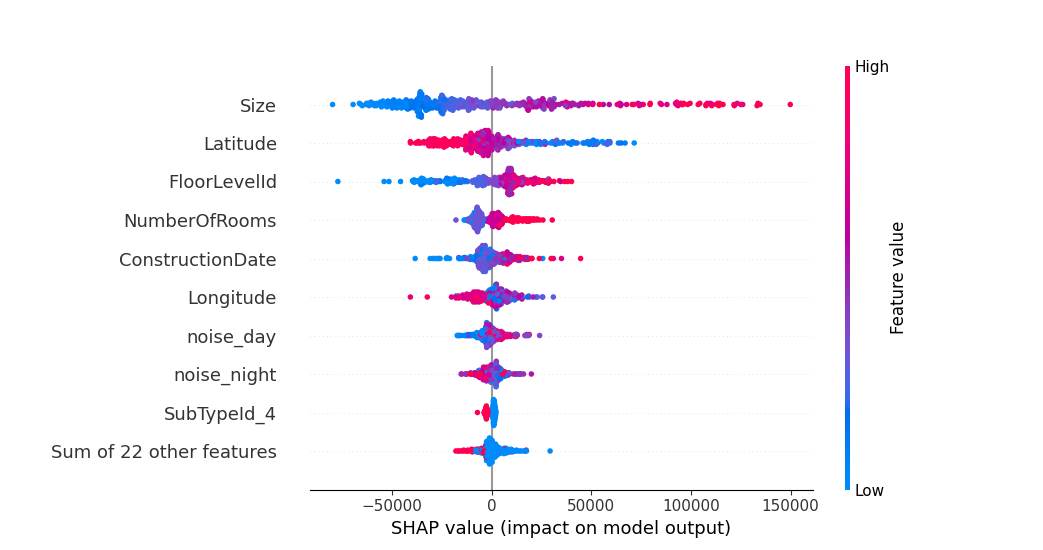}
\end{figure}

\begin{figure}[h!]
\caption{\csentence{LIME Plots for Area A}}\label{fig:r50A_lime}
\centering
\begin{subfigure}{.5\textwidth}
  \centering
  \includegraphics[width=1\linewidth]{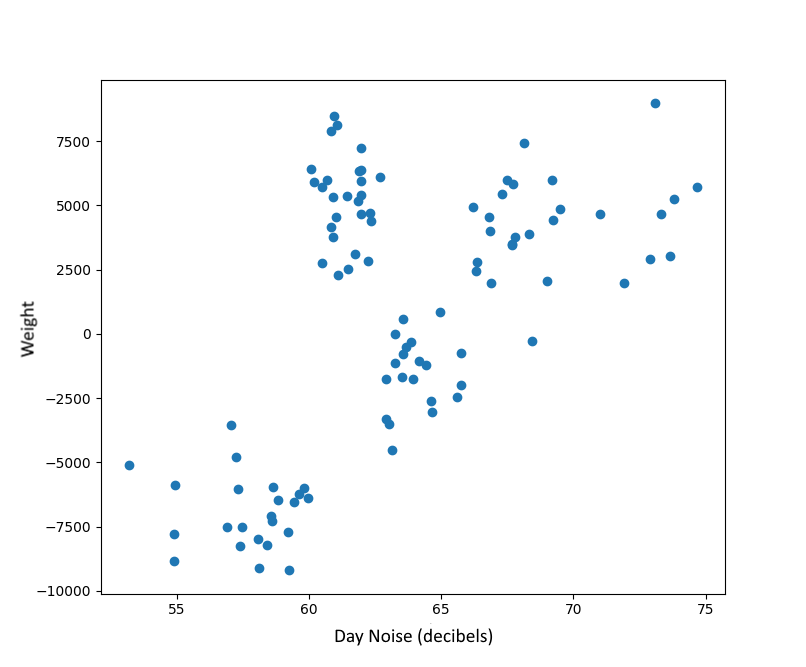}
  \caption{Day Noise}
  \label{fig:r50A_lime_day}
\end{subfigure}%
\begin{subfigure}{.5\textwidth}
  \centering
  \includegraphics[width=1\linewidth]{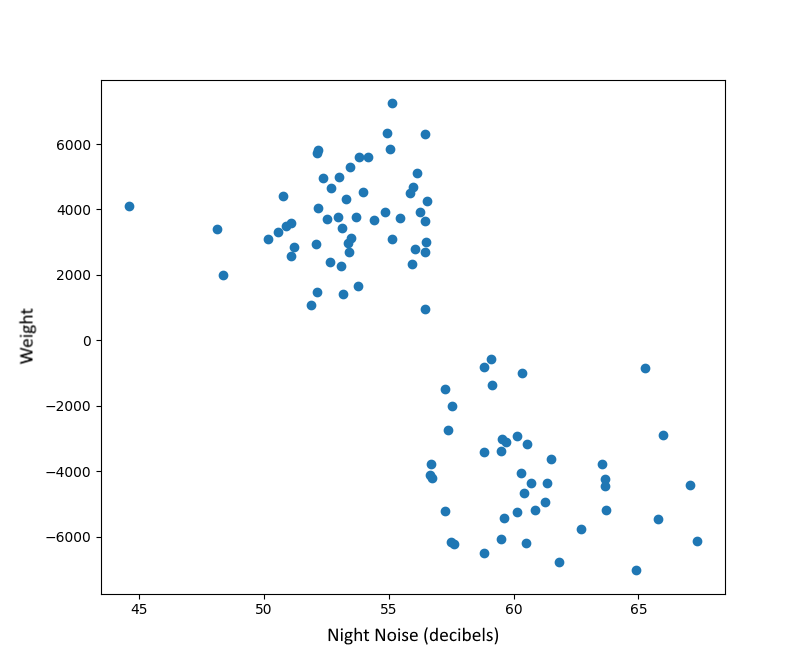}
  \caption{Night Noise}
  \label{fig:r50A_lime_night}
\end{subfigure}
\end{figure}

\begin{figure}[h!]
  \caption{\csentence{Feature Importance \& Permutation Importance for Area B}}\label{fig:r100B_featureImportance}
      \includegraphics[height=2.1in]{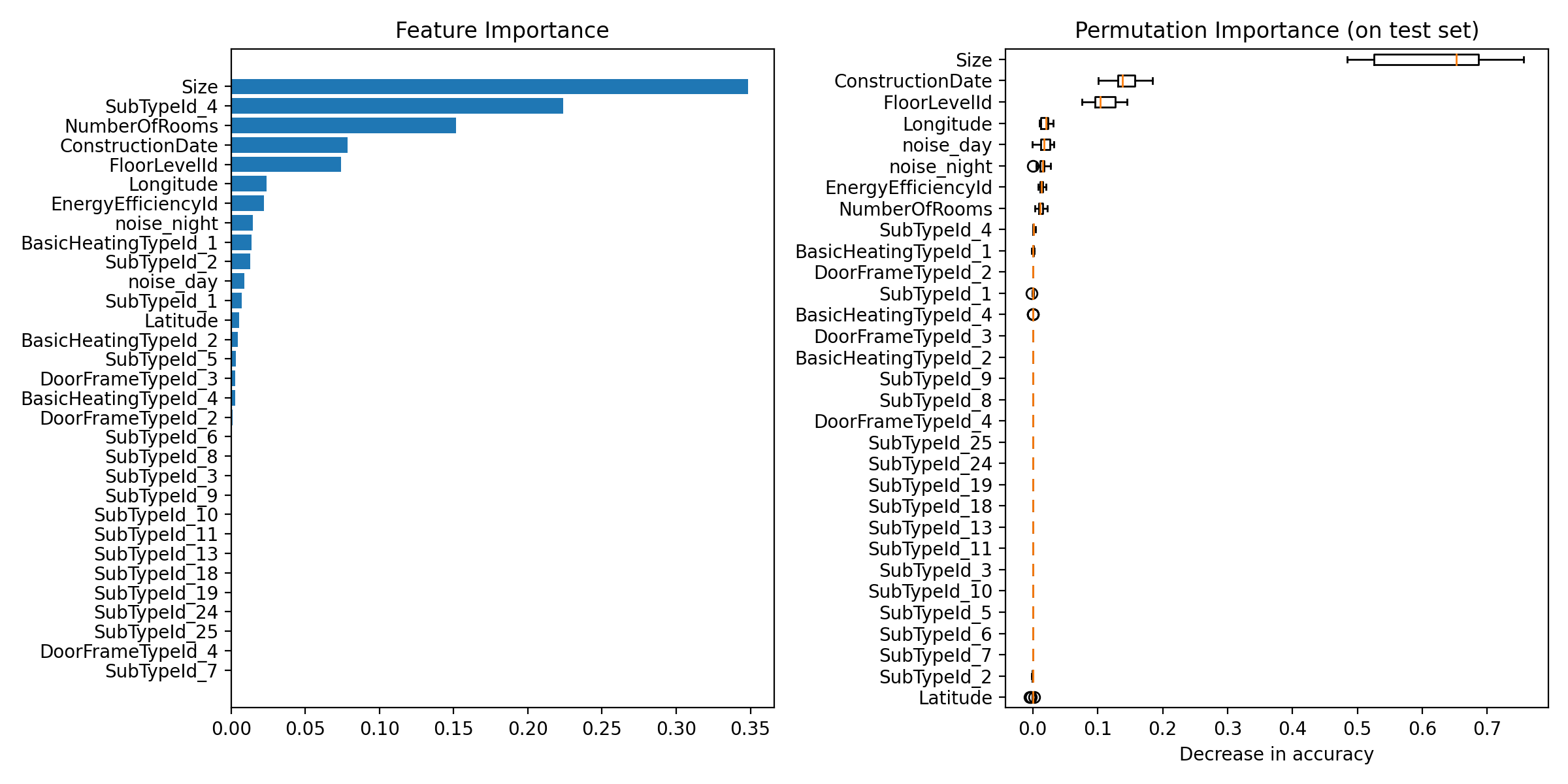}
\end{figure}

\begin{figure}[h!]
\caption{\csentence{Partial Dependence Plots for Area B}}\label{fig:r100A_PDP}
\centering
\begin{subfigure}{.5\textwidth}
  \centering
  \includegraphics[width=1\linewidth]{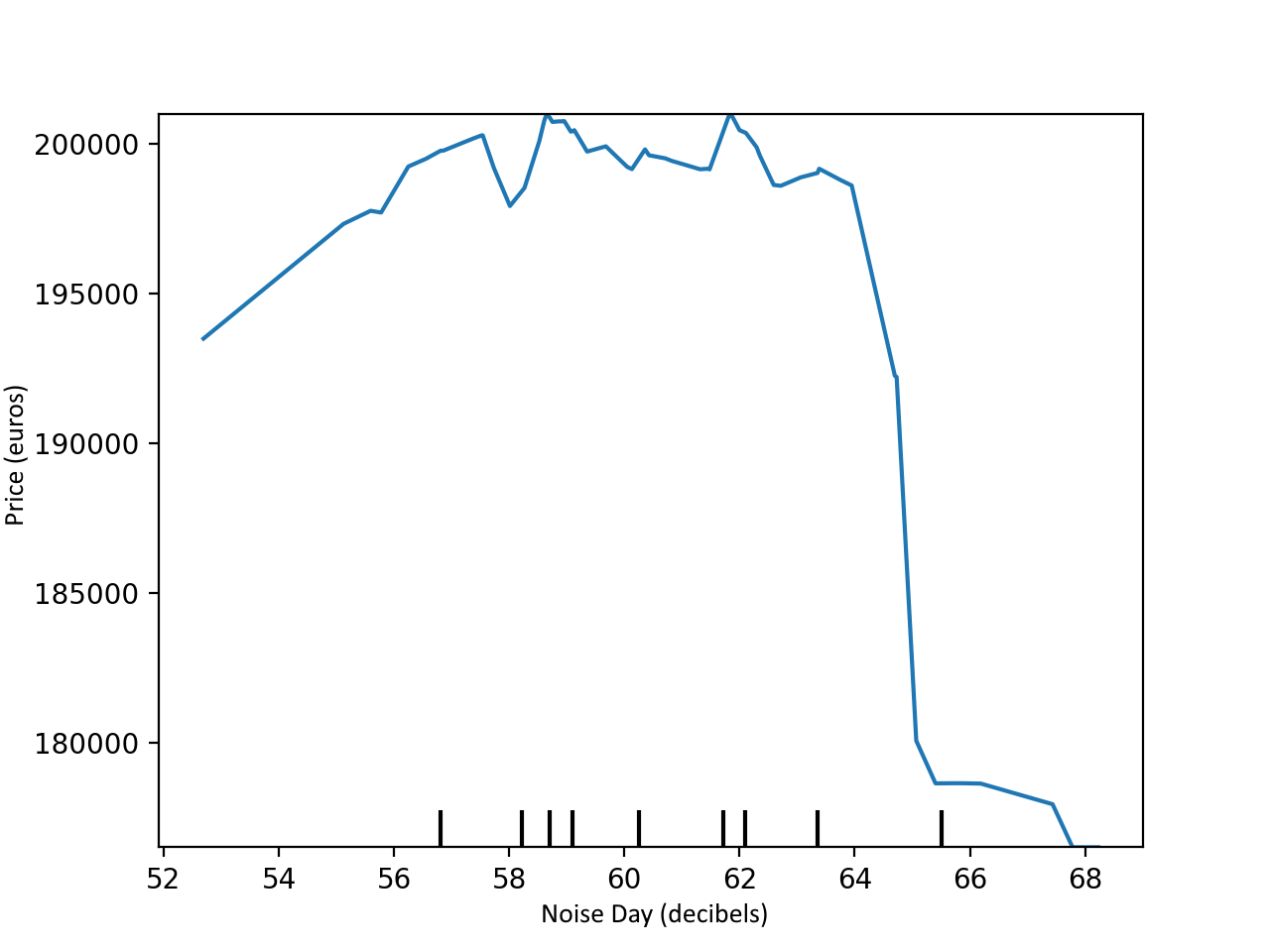}
  \caption{Day Noise}
  \label{fig:r100B_day}
\end{subfigure}%
\begin{subfigure}{.5\textwidth}
  \centering
  \includegraphics[width=1\linewidth]{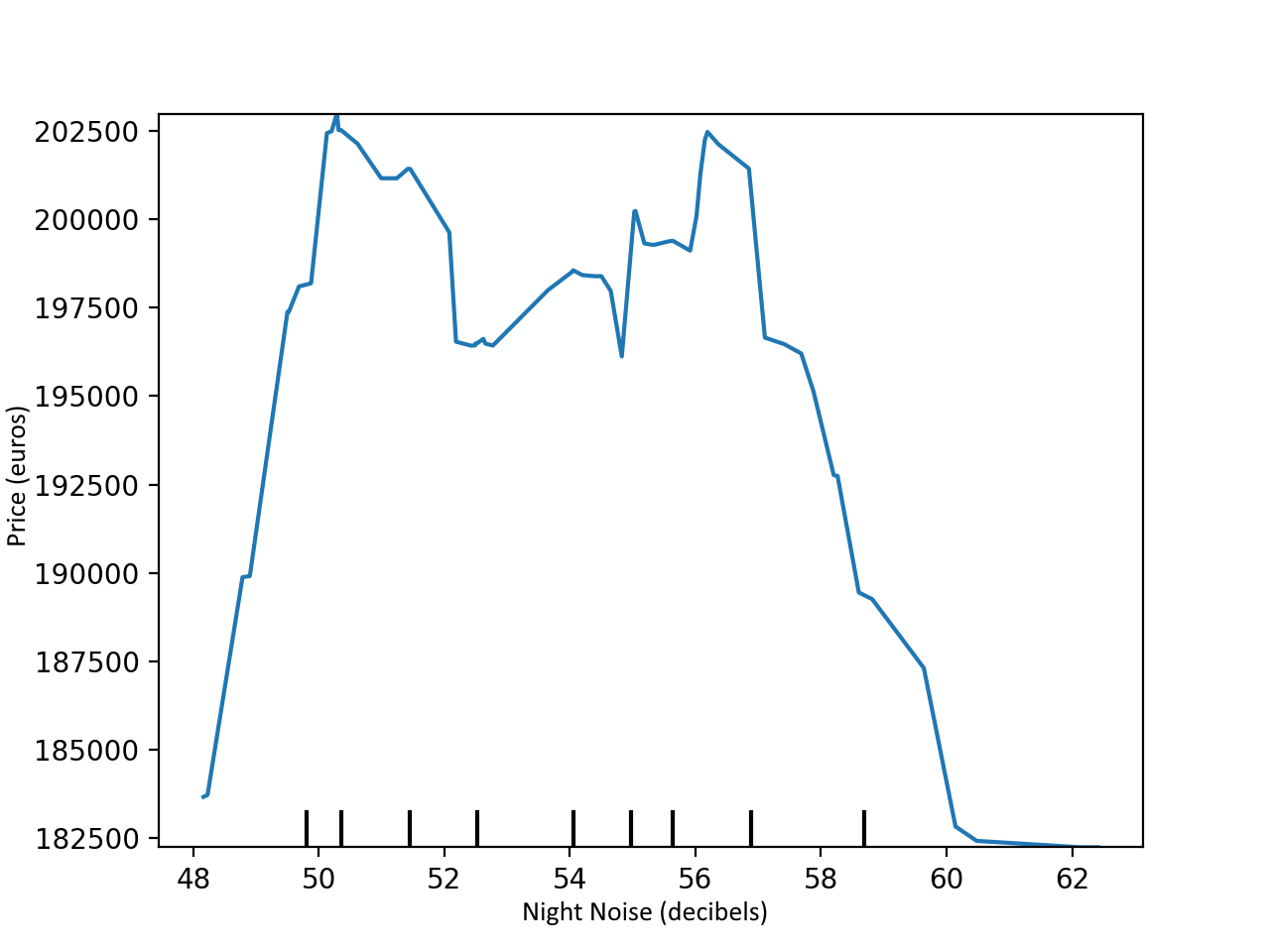}
  \caption{Night Noise}
  \label{fig:r100B_night}
\end{subfigure}
\end{figure}

\begin{figure}[h!]
  \caption{\csentence{SHAP Plot for Area B}}\label{fig:r100B_shap}
      \includegraphics[height=2.1in]{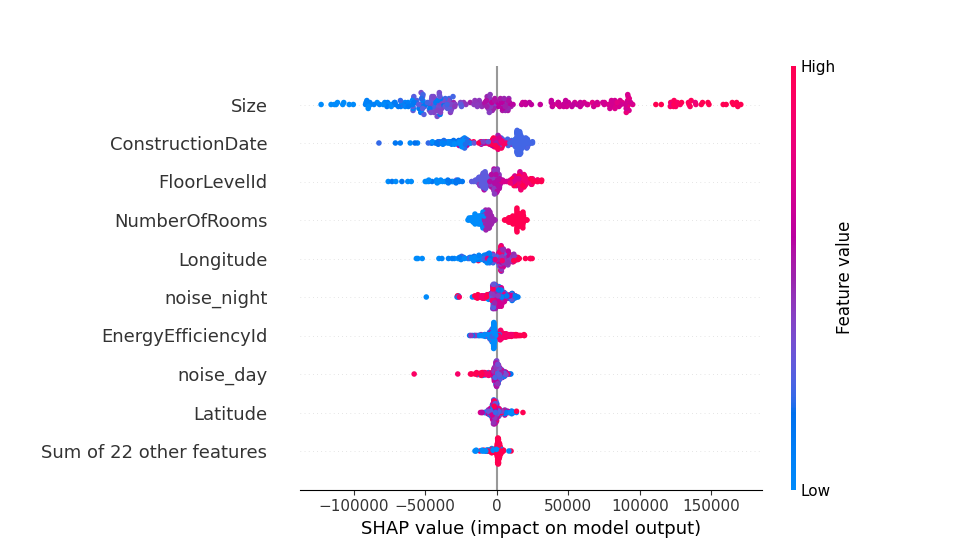}
\end{figure}

\begin{figure}[h!]
\caption{\csentence{LIME Plots for Area B}}\label{fig:r100B_lime}
\centering
\begin{subfigure}{.5\textwidth}
  \centering
  \includegraphics[width=1\linewidth]{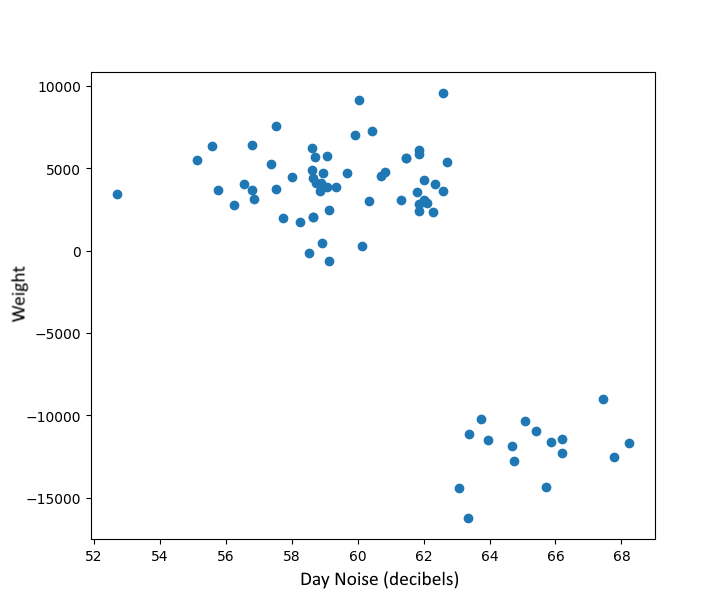}
  \caption{Day Noise}
  \label{fig:r100B_lime_day}
\end{subfigure}%
\begin{subfigure}{.5\textwidth}
  \centering
  \includegraphics[width=1\linewidth]{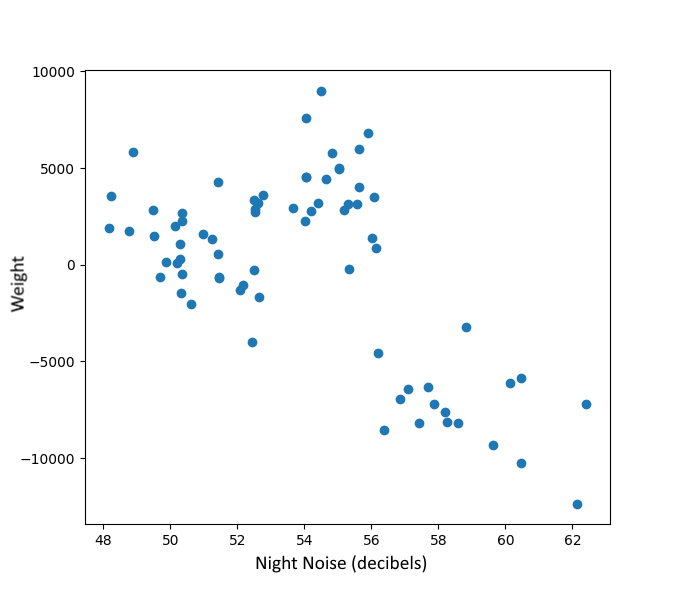}
  \caption{Night Noise}
  \label{fig:r100B_lime_night}
\end{subfigure}
\end{figure}

\begin{figure}[h!]
  \caption{\csentence{Feature Importance \& Permutation Importance for area C}}\label{fig:r50C_featureImportance}
      \includegraphics[height=2.1in]{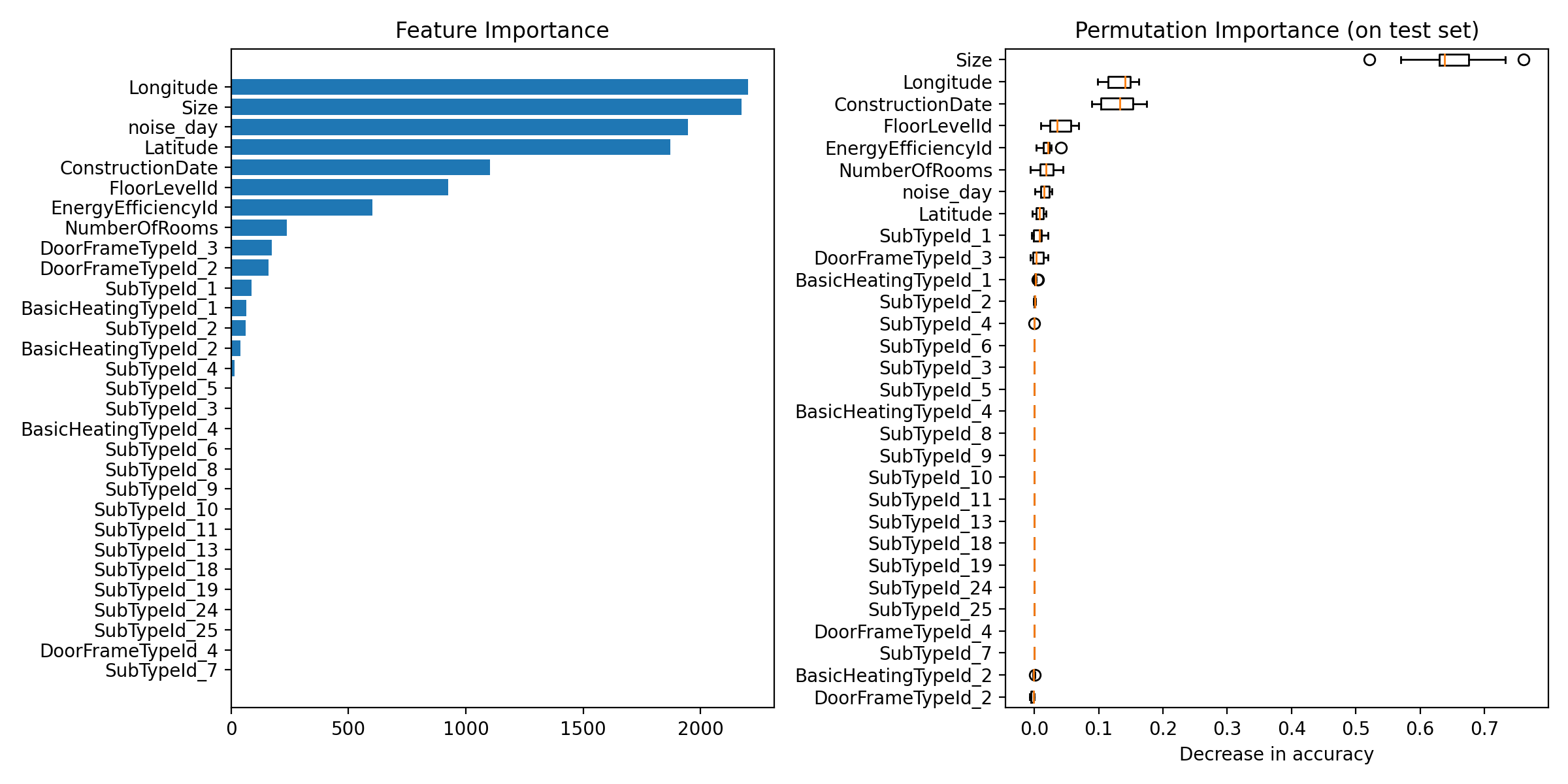}
\end{figure}

\clearpage

\begin{figure}[h!]
\caption{\csentence{Partial Dependence and LIME Plots for Area C}}\label{fig:r50C_PDP_lime}
\centering
\begin{subfigure}{.5\textwidth}
  \centering
  \includegraphics[width=1\linewidth]{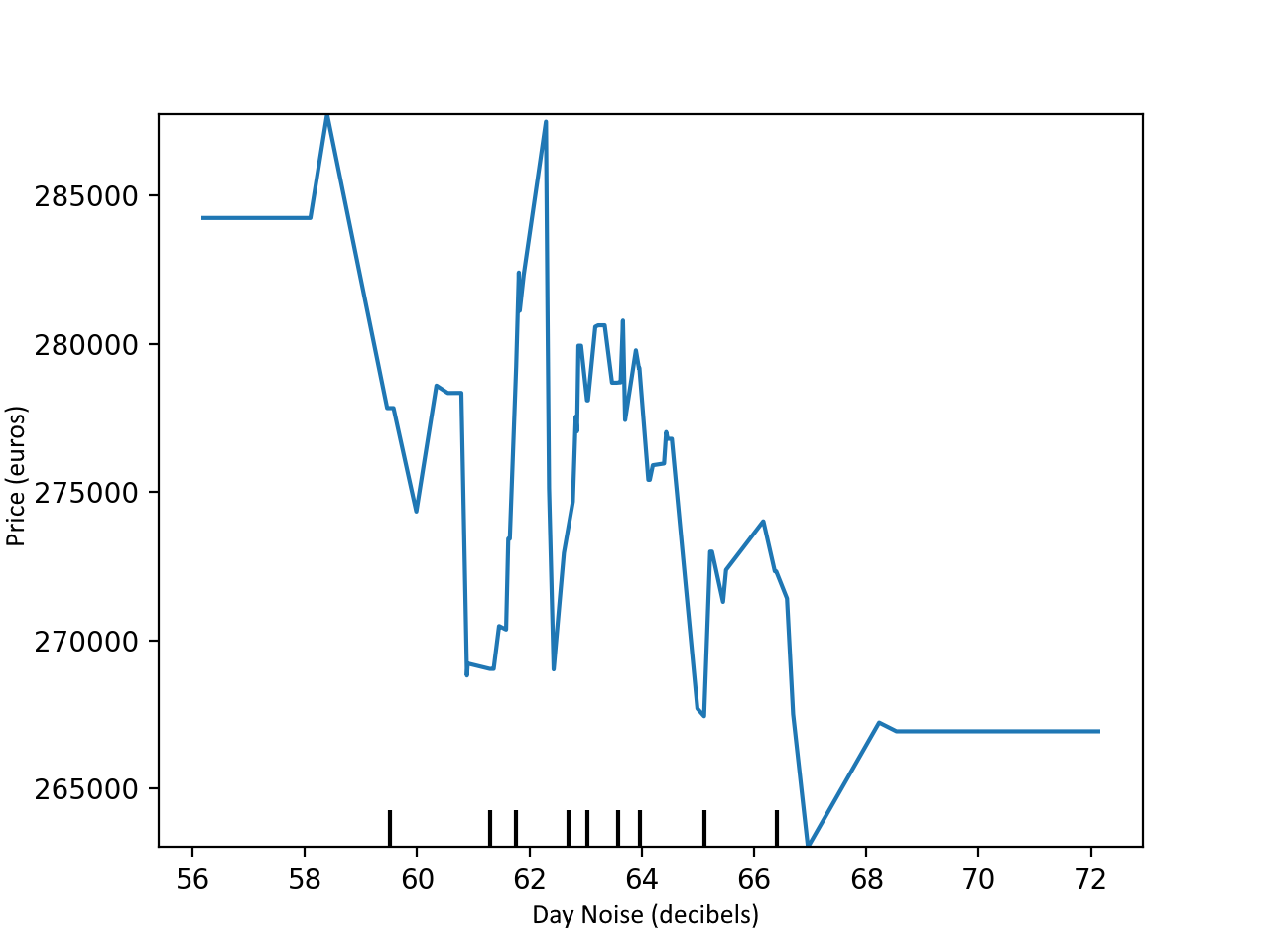}
  \caption{Partial Dependence (day noise)}
  \label{fig:r50C_day}
\end{subfigure}%
\begin{subfigure}{.5\textwidth}
  \centering
  \includegraphics[width=1\linewidth]{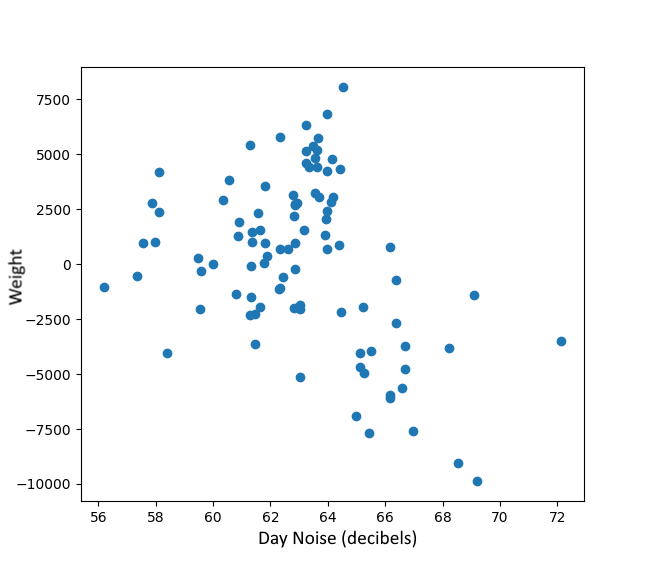}
  \caption{LIME Weights (day noise)}
  \label{fig:r50C_lime_day}
\end{subfigure}
\end{figure}

\begin{figure}[h!]
  \caption{\csentence{SHAP Plot for Area C}}\label{fig:r50C_shap}
      \includegraphics[height=2.1in]{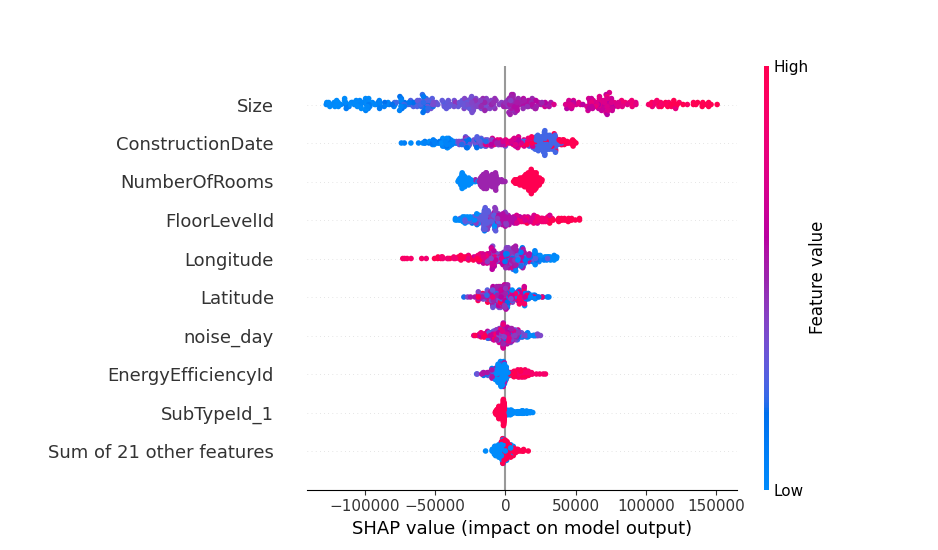}
\end{figure}


\begin{backmatter}

\section*{Acknowledgements}
  The authors thank to Anna-Maria Feneri for her helpful comments.

\section*{Abbreviations}
  DT, Decision Trees; GIS, Geographic Information System; IQR, Interquartile Range; LGBM, Light Gradient Boosting Models; LIME, Local Interpretable Model-Agnostic Explanations; MAPE, Mean Absolute Percentage Error; MAE, Mean Absolute Error; RF, Random Forest; SHAP, Shapley Additive Explanations; XGBoost, Extreme Gradient Boosting.

\section*{Competing interests}
  The authors declare that they have no competing interests.

\section*{Availability of data and materials}
  All data generated or analyzed during this study are included in this published article.

\section*{Authors' contributions}
    Conceptualization: GK, GT; Data collection: GK; Formal analysis: GK, Investigation: GK, Writing original manuscript: GK, GT, DV. All authors read and approved the final manuscript.


\bibliographystyle{bmc-mathphys} 
\bibliography{bmc_article}      

\end{backmatter}
\end{document}